\documentclass{article}
\usepackage{iclr2027_conference,times}   % NB: already loads xcolor
\usepackage{amsmath,amsfonts,bm}

\def\eqref#1{equation~\ref{#1}}
\def\1{\bm{1}}

\DeclareMathAlphabet{\mathsfit}{\encodingdefault}{\sfdefault}{m}{sl}
\SetMathAlphabet{\mathsfit}{bold}{\encodingdefault}{\sfdefault}{bx}{n}

\usepackage{hyperref}
\usepackage{url}
\usepackage{booktabs}
\usepackage{multirow}
\usepackage{graphicx}
\usepackage{amsmath,amssymb}
\usepackage{pifont}          % \ding{55} for the desiderata table

\makeatletter
\def\paragraph{\@startsection{paragraph}{4}{\z@}{0.6ex plus
0.2ex}{-1em}{\normalsize\bf}}
\makeatother
\usepackage{iftex}
\newif\ifcjk \cjkfalse
\ifXeTeX \cjktrue \fi
\ifLuaTeX \cjktrue \fi
\ifcjk
  \usepackage{fontspec}
  \IfFontExistsTF{TeX Gyre Termes X}{%
    \setmainfont{TeX Gyre Termes X}%
    \setsansfont{TeX Gyre Heros}[Scale=MatchLowercase]%
    \setmonofont{TeX Gyre Cursor}[Scale=MatchLowercase]%
  }{%
    \IfFontExistsTF{Times New Roman}{%
      \setmainfont{Times New Roman}%
      \IfFontExistsTF{Helvetica}{\setsansfont{Helvetica}[Scale=MatchLowercase]}{}%
      \IfFontExistsTF{Courier New}{\setmonofont{Courier New}[Scale=MatchLowercase]}{}%
    }{}%
  }%
  \IfFontExistsTF{Songti SC}{%
    \newfontfamily\cjkfont{Songti SC}[Scale=0.95]%
  }{%
    \IfFontExistsTF{Noto Serif CJK SC}{%
      \newfontfamily\cjkfont{Noto Serif CJK SC}[Scale=0.95]%
    }{%
      \IfFontExistsTF{PingFang SC}{%
        \newfontfamily\cjkfont{PingFang SC}[Scale=0.95]%
      }{%
        \cjkfalse
      }%
    }%
  }
\fi
\ifcjk
  \newcommand{\zh}[1]{{\cjkfont #1}}
  \newcommand{\todofont}{\cjkfont}
\else
  \usepackage{CJKutf8}
  \newcommand{\zh}[1]{\begin{CJK}{UTF8}{gbsn}#1\end{CJK}}
  \newcommand{\todofont}{}
\fi

\definecolor{todored}{rgb}{0.75,0.12,0.10}
\definecolor{numblue}{rgb}{0.10,0.30,0.70}
\newif\ifshowtodo \showtodofalse
\ifshowtodo
  \newcommand{\todo}[1]{{\color{todored}\textbf{[TODO:~}{\todofont #1}\textbf{]}}}
\else
  \newcommand{\todo}[1]{}
\fi
\newif\ifhighlightnum \highlightnumfalse
\ifhighlightnum
  \newcommand{\num}[1]{\textcolor{numblue}{#1}}
\else
  \newcommand{\num}[1]{#1}
\fi

\newcommand{\figslot}[3]{%
  \setlength{\fboxrule}{0.6pt}%
  \fcolorbox{todored}{white}{%
    \begin{minipage}[c][#2][c]{\dimexpr#1-2\fboxsep-2\fboxrule\relax}%
      \centering\color{todored}\footnotesize
      \textbf{[FIGURE SLOT]}\quad(\the\dimexpr#1\relax\ $\times$ \the\dimexpr#2\relax)\\[3pt]
      {\todofont #3}%
    \end{minipage}}}

\newcommand{\st}{\emph{Squint Text}}

\newcommand{\stimgs}{squint-text images}
\newcommand{\base}{\textsc{Base}}
\newcommand{\ours}{\textsc{Squint~OPSD}}
\newcommand{\bench}{\textsc{SquintBench}}
\newcommand{\probeset}{\textsc{SquintCtrl}}
\newcommand{\ctrlpair}{\textsc{SquintCtrl-Pair}}      % 500 pairs, presence
\newcommand{\ctrlglyph}{\textsc{SquintCtrl-Glyph}}    % 2x500, closed-set identity
\newcommand{\ctrlstr}{\textsc{SquintCtrl-Strength}}   % 200 scenes x 5 strengths
\newcommand{\ctrlread}{\textsc{SquintCtrl-Read}}      % 200 random glyph x scene
\newcommand{\amr}{\mathrm{AMR}}

\colorlet{blue}{black}
\colorlet{numblue}{black}

\title{When MLLMs Miss What Humans Read: \\
Diagnosing and Learning to Read \\
Squint Text}

\author{%
Chen Peng$^{1}$, Shan Huang$^{2}$, Huang Jianbo$^{2}$, Linrui Qin$^{2}$, Qibo Sun$^{1}$, Mengwei Xu$^{1}$ \\
$^{1}$State Key Laboratory of Networking and Switching Technology, \\
Beijing University of Posts and Telecommunications, Beijing, China \\
$^{2}$Independent Researchers
}

\iclrfinalcopy % arXiv: show authors and suppress review line numbers.
\begin{document}
\maketitle
\lhead{} % arXiv preprint: suppress the conference-publication running head.

\begin{abstract}
Multimodal LLMs can fail to read text embedded in an image's coarse
structure---text a human recovers by squinting---and report that no text is
present. We name this class \st{} and study it as a recognition deficiency,
motivated by cases from a production moderation pipeline. We introduce
\bench{}, comprising \num{1076} production-derived moderation images, and
\probeset{}, a controlled synthetic suite for mechanism diagnosis. Frozen-weight
analysis shows that hidden text leaves a coarse signal while glyph identity is
\textcolor{blue}{weakly decodable under global pooling. Restricting read-out to the text region improves glyph-identity decoding, identifying spatial selection as an important factor in glyph-identity read-out; an exact glyph mask further improves decoding.} Suppressing high-frequency content also improves reading at a fixed
visual-token budget, while visual token count may also contribute. These findings motivate \ours{}: a frozen teacher receives a
downscaled view and the ground-truth transcription as privileged context, and
its token-level distributions supervise a student along on-policy trajectories.
At inference, the student reads the original image alone. \ours{} reaches
\num{88.61} ANLS on \bench{}, versus \num{26.50} for the unadapted backbone,
\num{24.44} for the best general-purpose model on the original view, and
\num{68.60} at best for an off-the-shelf model under a downscaled view. These
results substantially narrow the gap between available visual evidence and its
generated read-out without auxiliary views at inference.
\end{abstract}
      % budget  170 w
% =====================================================================
%  1  INTRODUCTION
%  Revised structure: overview -> problem -> deployment phenomenon
%  -> diagnosis/method -> results -> contributions.
% =====================================================================
\section{Introduction}
\label{sec:intro}

Multimodal LLMs can miss text embedded in an image's coarse structure even
when a human can recover it by squinting or viewing the image at reduced scale.
We call this failure \textbf{\st{}}; we study why the visual evidence is
difficult for current models to access and show that the corrective viewing
condition can be internalized through training, enabling original-only
inference.

\begin{figure}[t]
\centering
\includegraphics[width=\textwidth]{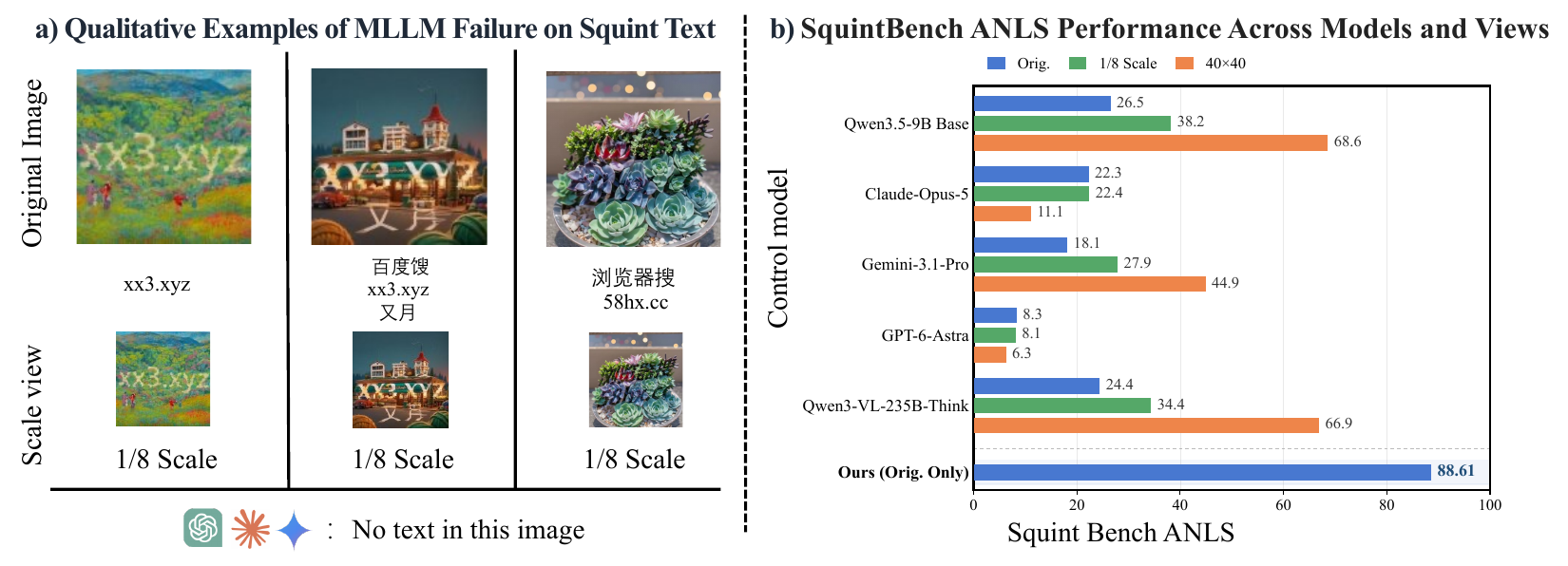}
\caption{\textbf{\st{}, and why the deployed defence does not remove it.}
\textbf{(a)}~Pipeline images whose embedded string a human reads---directly or
by stepping back---while GPT, Gemini and Claude each answer \emph{``No text in
this image''}, on the original and the scaled view alike.
\textbf{(b)}~Downscaling before review recovers only part of the hidden text
for every off-the-shelf model.}
\label{fig:teaser}
\end{figure}

Modern VLMs are strong scene-text readers, yet this capability breaks sharply
on \st{}. As Figure~\ref{fig:teaser} shows, frontier commercial and open-source
models \citep{singh2026openaigpt5card,bai2025qwen3vltechnicalreport} can report that no text is present
even when the hidden string is readily recoverable by a person. The failure also persists with model scale
(Table~\ref{tab:mllm}). Downscaling or low-pass filtering recovers part of the
missing text, but leaves a substantial residual. \st{} therefore exposes a
specific recognition deficiency that stronger generic perception and
inference-time preprocessing do not reliably remove.

This deficiency matters in deployment. Images are widely used to carry illicit
promotion and other policy-violating content
\citep{metarestricted,metahateful,xillegal,tiktokrca}, and the harmful
semantics can reside in text drawn into the image rather than in the depicted
scene \citep{299571,NEURIPS2020_1b84c4ce,wu2025icmassistantinstructiontuningmultimodallarge}. In production moderation
traffic, we observe domains, contact strings, and slogans embedded into coarse
image structure so that human viewers can read them while the reviewing model
reports no text. Such examples create a direct blind spot at the visual-review
stage of an automated moderation pipeline \citep{metacser}.

To study the failure, we introduce \bench{}, containing \num{1076} production
moderation images, and \probeset{}, a controlled synthetic suite for isolating
what the model represents and where the failure arises. Frozen-weight
diagnostics show that hidden text leaves a strong coarse signal, while glyph
identity is only weakly represented; identity becomes substantially more
accessible when the correct spatial regions are supplied. Controlled image
interventions on \bench{} itself further show that suppressing
high-frequency content can substantially improve reading at matched
visual-token budgets, implicating high-frequency interference as a major
contributor while leaving room for a secondary token-count effect. These findings motivate \ours{}: during training,
a frozen teacher receives a downscaled view and the ground-truth transcription
as privileged context, and its token-level distribution supervises a student
that sees only the original image. At inference, the student receives the
original image alone.

\ours{} reaches \num{88.61} ANLS on \bench{}, compared with \num{26.50} for the
base model and \num{58.96} when the corrective view is supplied only at
inference. The capability transfers to external hidden-text benchmarks while
retaining substantial general vision-language capability with measurable,
task-dependent trade-offs. Analysis after training shows that glyph
representations become more discriminative, reading depends causally on
glyph-bearing regions under our visual-token embedding intervention, and the residual
candidate distribution becomes more visually structured. Code and the
evaluation suite are released at
\url{https://anonymous.4open.science/r/Squint_Text-3456}.

\paragraph{Contributions.}
\begin{itemize}\itemsep1pt \parskip0pt
\item \textbf{A deployment-grounded failure mode and two complementary
datasets.} We define \st{} and introduce \bench{}, comprising \num{1076}
production moderation images, for realistic evaluation, together with
\probeset{} for controlled diagnosis of hidden-text representation,
localization, and recoverability.
\item \textbf{A diagnosis of why current MLLMs miss \st{}.} Hidden text leaves
a strong coarse signal but weak glyph identity; the identity signal becomes
more accessible at glyph-bearing locations, and high-frequency interference is
a major contributor, with token count a possible secondary factor.
\item \textbf{A diagnosis-driven training method.} \ours{} uses a downscaled
privileged view together with a ground-truth hint to provide token-level
distributional supervision for original-only inference, reaching \num{88.61}
ANLS on \bench{}.
\end{itemize}

% ---- Figure 1: TEASER --------------------------------------------------
         % budget  900 w   + Fig 1 (related work folded in)
% =====================================================================
%  2  PRELIMINARIES AND DATA
% =====================================================================
\section{Preliminaries and Data}
\label{sec:prelim}

\subsection{Problem setting}
\label{sec:setting}
Given an image $x$ and an instruction to transcribe its text, the model must
report every string a human reader would, including text carried in the image's
coarse structure. We use \emph{read-out} to refer to the mapping from encoded
visual representations to the generated transcription. Let $v=g(x)$ denote a
bicubically downscaled view of $x$, at $1/8$ linear scale unless stated
otherwise. We compare a single-image input $x$ with a two-view input
$\langle x,v\rangle$ to test whether changing the viewing condition makes the
same underlying visual evidence easier for a frozen model to use. Section~\ref{sec:diagnosis}
performs this comparison with identical off-the-shelf weights; the training
method is introduced separately in Section~\ref{sec:method}.

On \bench{} we report ANLS \citep{Biten_2019_ICCV}, with predicted and reference
regions matched as a set before scoring so that a correct transcription written
in a different order is not penalised.
\todo{确认 set-matched 是最终口径，并在附录报它与顺序敏感变体的一致性。}

\subsection{Datasets}
\label{sec:data}
\label{sec:bench}
\label{sec:probeset}
Showing that the phenomenon occurs in deployment requires real model failures;
attributing the failure to a particular mechanism requires controlled examples.
We therefore use two complementary datasets.

\bench{} comprises \num{1076} \stimgs{} collected from a production moderation
pipeline. All \num{1076} examples are production-derived. These images are typically profile photos.
\textcolor{blue}{Six participants transcribed all \num{1076} \bench{} images, achieving an ANLS of 98.6\%.} Human annotators transcribe all visible and hidden text. \textcolor{blue}{The transcriptions were checked by multiple annotators.} Thus,
\bench{} is used as the realistic end-to-end evaluation set.

\textcolor{blue}{Of the \num{1076} images, \num{1035} contain both Han characters and Latin letters. Within these mixed-script images, the ground-truth transcriptions contain \num{4588} Han characters and \num{4303} Latin letters, an approximately $1{:}1$ ratio.}
\textcolor{blue}{It is constructed separately from the synthetic training data; the training sources and their relation to \probeset{} are detailed in Appendix~\ref{app:trainmix}.}
\todo{补采样口径（时间窗、去重、标注一致性）与数据治理说明。}

\probeset{} supplies the controlled counterpart. A glyph string is rendered as
a binary conditioning image and embedded into a scene using QR Code Monster v2
\citep{qrmonster}, a ControlNet \citep{Zhang_2023_ICCV} over Stable
Diffusion~1.5 \citep{Rombach_2022_CVPR}, with scene prompts sampled from JourneyDB
\citep{DBLP:conf/nips/SunPGLDWZZQWDQW23}. For each controlled comparison, non-target generation
variables are held fixed; because the conditioning image is known, each sample
also has an exact token-level glyph mask without detector or human annotation.

The four subsets serve distinct diagnostics. \emph{\ctrlpair{}} contains
\num{500} matched text-present/text-free pairs for testing whether hidden-text
presence is encoded. The text-free member is regenerated through the same
ControlNet branch using an all-black conditioning image, avoiding a synthesis
pipeline shortcut. \emph{\ctrlglyph{}} contains two \num{500}-image closed
vocabularies over ten Chinese glyphs
(\zh{加}\zh{女}\zh{可}\zh{度}\zh{搜}\zh{月}\zh{百}\zh{看}\zh{网}\zh{馊}) and ten
Latin letters for glyph-identity and localization analyses. \emph{\ctrlstr{}}
varies \num{200} scenes across \num{5} conditioning strengths to study the
recoverability boundary. \emph{\ctrlread{}} contains \num{200} open-vocabulary
images for end-to-end read-out and post-training analysis. Detailed generation
and control protocols are provided in Appendix~\ref{app:synthesis}, and
representative examples of each subset are shown in
Figure~\ref{fig:ctrl-examples}.
\section{Diagnosing Squint Text}
\label{sec:diagnosis}

Unless stated otherwise, all diagnostic experiments use Qwen3.5-9B
\citep{qwen35} with frozen weights. We report set-matched ANLS for transcription and out-of-fold AUC for
probing. The failure is not confined to this checkpoint: four commercial and
open-weight MLLMs obtain only \num{8.3}--\num{24.4} ANLS on the original
\bench{} images
(Table~\ref{tab:mllm}).
\todo{复现性：在另外 $N$ 个开源、$M$ 个闭源模型上复现本节三个结论；届时再新建 Cross-model Replication 附录并在此引用。}

\subsection{Hidden-Text Representation}
\label{sec:encoded}
To test what survives in the vision encoder, we fit logistic-regression probes
to frozen, pre-merger patch features at blocks L6/12/18/24/27. Each probe uses
mean-pooled features and is evaluated out of fold against shuffled-label
controls; fitting it does not update the MLLM. On \ctrlpair{}, whose \num{500}
pairs differ only in whether a string was embedded, text presence reaches
\num{0.956}--\num{0.981} AUC. Label shuffling returns chance-level performance
(\num{0.47}--\num{0.52} AUC), while matched generation paths control for
synthesis-pipeline artefacts (Appendix~\ref{app:fingerprint}).
Going beyond mere presence, we further test how well each individual glyph can
be read out on the closed-vocabulary \ctrlglyph{} set. Glyph identity is
weaker than presence but remains above chance: Chinese macro-AUC peaks at
\num{0.718} at L18 and averages \num{0.676} across the five blocks, with
Latin at \num{0.612} and \num{0.581}, respectively. Thus, the encoder detects
hidden text, although glyph identity is only weakly represented.

% \subsection{Glyph identity is accessible at the right locations}
% \label{sec:localization}
% %  CORE RESULT.  The RANDOM condition is what makes this a proof rather
% %  than an observation.  Do not let it get compressed away.
% For the $1024^2$ probe input, the vision tower uses $16{\times}16$ patches,
% yielding a $64{\times}64$ grid of \num{4096} pre-merger features. We align the
% known \ctrlglyph{} mask to this grid and compare three poolings of the same
% features: \textsc{oracle} averages glyph-bearing patches, \textsc{global}
% averages the full grid, and \textsc{random} averages an equally sized random
% subset. \textsc{Random} controls for sample count: it estimates the global mean
% and empirically differs from it by at most \num{0.004} AUC. Yet
% Figure~\ref{fig:diagnosis}a shows
% $\textsc{random}\approx\textsc{global}\ll\textsc{oracle}$, with \textsc{oracle}
% ahead by ${\approx}\num{0.1}$ AUC. Glyph identity is therefore substantially
% more accessible at the correct locations, identifying spatial selection as a
% major bottleneck. The $2{\times}2$ patch merger changes separability by at most
% \num{0.018} AUC (Appendix~\ref{app:layers}).
\subsection{Spatial Localization}
\label{sec:localization}
\begingroup\color{blue}
We compare mean-pooling rules on the same frozen, pre-merger visual features,
then train linear probes for glyph identity; no MLLM weights are updated.
\textsc{Oracle} pools the $K$ glyph-bearing tokens selected by the exact mask,
whereas \textsc{global} pools all tokens and \textsc{global random} pools $K$
random tokens. Because the exact mask specifies both location and glyph shape,
its advantage alone does not isolate localization.
We therefore add two controls. \textsc{Translated mask} shifts the same mask
by a nonzero displacement, preserving its shape and $K$ without clipping or
wrapping. \textsc{Local random K} samples $K$ tokens from the coarse window
containing the text, without following its strokes. Both controls use five
random draws per image.
Figure~\ref{fig:diagnosis}a shows that local random pooling exceeds global
random pooling across the tested layers. At L18, macro-AUC is \num{0.774}
for oracle, \num{0.684} for local random, and \num{0.649} for translated masks.
These results indicate that spatial selection is an important factor in reading out hidden-text information: pooling within the text region improves glyph recognition, and precise glyph-aligned selection further improves it
\citep{Wu_2024_CVPR,zhang2025mllms}.
Passing these features through the patch merger changes separability by at
most \num{0.018} AUC (Appendix~\ref{app:layers}).
\endgroup

\subsection{Frequency and Token Count}
\label{sec:freq}
%  !! POSITIONING -- do not undo !!
%  That downscaling / low-pass RECOVERS hidden text is already published
%  (SemVink, Illusory VQA).  Our contribution is the SEPARATION of
%  mechanisms.  The opening concession is what makes the rest defensible.
%
%  REVISED 2026-09-21: neighbourhood averaging also attenuates high
%  frequencies, so the current experiment does not independently identify
%  a token-count effect.  Do not describe it as a token-only control.
That downscaling or low-pass filtering makes such text legible is already
established---HC-Bench evaluates after downscaling \citep{li-etal-2025-semvink} and Illusory
VQA after low-pass filtering \citep{11147971}. What is not established is
\emph{which} property of those transforms is responsible, and the answer
determines what one should build: (\textbf{H1}) high-frequency texture masks the
glyph structure; (\textbf{H2}) too many tokens dilute attention; (\textbf{H3}) it
is a contrast problem.

\begin{figure}[!tb]
\centering
\includegraphics[width=\textwidth]{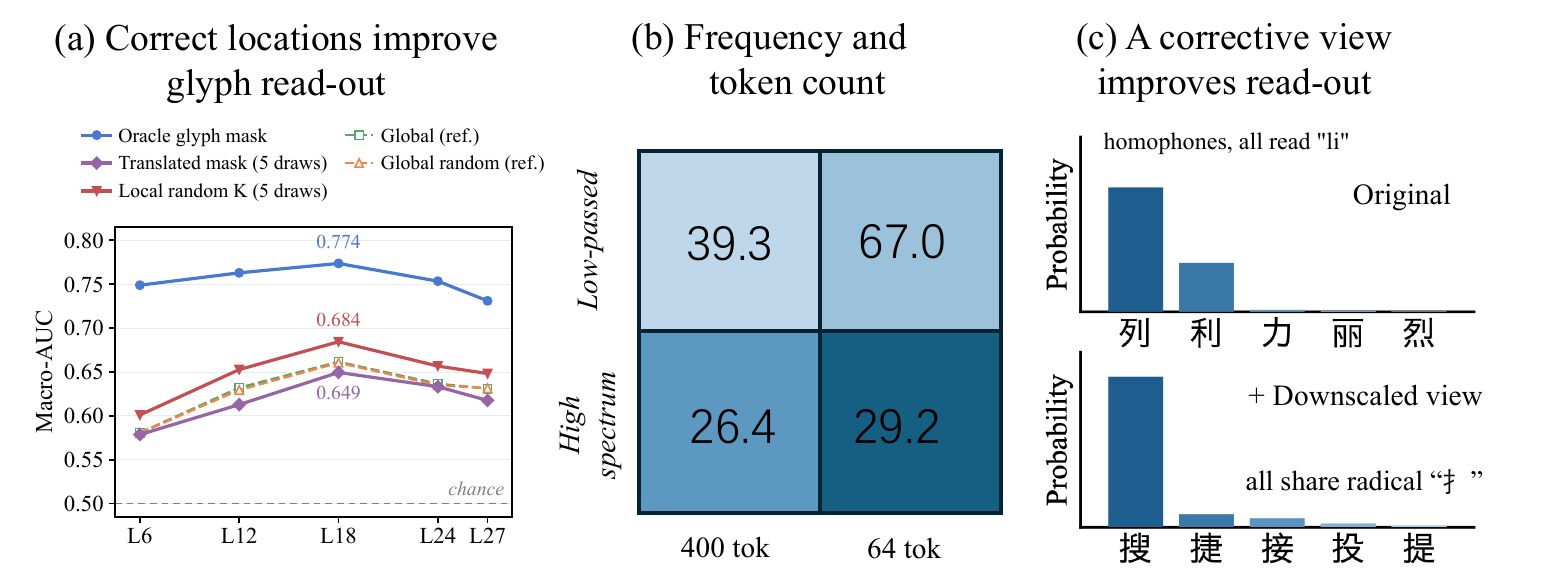}
\caption{\textbf{Three diagnostics on frozen off-the-shelf weights.}
\textcolor{blue}{\textbf{(a)}~Linear-probe glyph-identity macro-AUC on frozen visual features. Translated masks preserve mask shape and token count under nonzero shifts without clipping or wrapping; local random pooling selects the same number of tokens within the coarse text window (five draws each). Coarse location improves read-out, with an additional gain from exact glyph-aligned selection.}
\textbf{(b)}~At a fixed \num{64}-token budget, suppressing high-frequency content
raises squint-text ANLS from \num{29.2} to \num{67.0} (per-$\sigma$
results in Appendix~\ref{app:freq}).
\textbf{(c)}~For the same target glyph \zh{搜}, adding a $1/8$-scale view of
the same image shifts the base model's phonetic predictions toward visually matching glyphs, indicating stronger visual
grounding.}
\label{fig:diagnosis}
\end{figure}

We test these explanations on \bench{} images
(Figure~\ref{fig:diagnosis}b). At the original token budget (median \num{400}),
Gaussian low-pass filtering raises ANLS from \num{26.4} to \num{39.3},
showing that suppressing high frequencies helps without shortening the visual
sequence. Restricting the visual input to \num{64} tokens yields \num{29.2};
applying a stronger low-pass transform at the same token budget raises ANLS
to \num{67.0}. This matched-budget comparison independently supports H1.
% However, constructing the \num{64}-token input also changes its frequency
% content, so it does not isolate H2. Contrast-only transforms provide no such gain
% (Appendix~\ref{app:freq}), ruling out H3 as the main explanation. These results
% identify high-frequency interference as a primary barrier, but do not establish
% whether reducing token count provides an additional benefit.
However, constructing the \num{64}-token input also alters its frequency content, 
so H2 cannot be fully disentangled from H1. We additionally test H3 with 
contrast-only transforms, which yield substantially smaller gains than frequency suppression (Appendix~\ref{app:freq}), 
making contrast an unlikely primary explanation. 
Overall, these results identify high-frequency interference as a major contributor,
while visual token count may also contribute, although its effect is not fully disentangled from frequency.

% ---- Figure 2: THE MOTIVATION (three-panel diagnosis chain) -------------
%  Three panels that read left to right as one causal chain:
%   (a) glyph read-out improves at the correct locations
%   (b) frequency removal helps; the token-count effect remains unresolved
%   (c) remove the frequency and the model reads -- with no new pixels
%  !! Do NOT add SFT / OPSD rows to panel (c): neither method exists yet
%     at this point in the paper.

\subsection{Corrective Viewing}
\label{sec:bridge}
%  BRIDGE.  Still base weights, still only the input changes.  This is
%  what makes Section 4 look derived rather than proposed.
We next give the same frozen model the two-view input $\langle x,v\rangle$, with
no change to its weights or prompt. On \bench{}, adding the $1/8$-scale view
raises ANLS from \num{26.50} to \num{58.96}. Figure~\ref{fig:diagnosis}c shows the
corresponding shift from phonetic alternatives toward candidates that share the
target's visual structure. Because $v=g(x)$ is deterministic and lossy, the gain is consistent with the
same frozen weights exploiting coarse visual evidence more effectively under a
different viewing condition. The result motivates learning a similar read-out
behaviour while retaining original-only input.

%  !! This is the ONLY explanation of D1/D2/D3 in the paper.  Table 1's
%  caption and Sec.4's prose both refer back here and must not restate.
\noindent\textbf{Design requirements.}\ The diagnosis yields three requirements.
\textbf{(D1) Original-only inference:} the solution should receive $x$ alone,
because the needed evidence already survives in its representation.
\textbf{(D2) Localization-bearing supervision:} a transcript says what the
glyph is but not where its evidence lies, whereas predictions elicited by the
downscaled view reflect evidence that has become easier to select. The method
should transfer this implicit spatial information; whether it does so is tested
in \S\ref{sec:amr}.
\textbf{(D3) Distributional targets:} at each step the teacher assigns
probability mass over possible next tokens, not only the ground-truth one;
the relative probabilities of alternatives retain which glyphs remain
visually plausible --- information a one-hot label cannot express.

\begin{table}[!tb]
\centering\small
\caption{Candidate solutions under the requirements of
\S\ref{sec:bridge}: \textbf{D1} uses original-only inference, \textbf{D2} uses
localization-bearing supervision, and \textbf{D3} uses a distributional target.
A dash denotes a method with no training target. Only \ours{} satisfies all
three.}
\label{tab:desiderata}
\begin{tabular}{@{}lccc@{}}
\toprule
& D1 & D2 & D3 \\
\midrule
ground-truth SFT                                  & \checkmark & \ding{55} & \ding{55} \\
sparse-reward RL                                  & \checkmark & \ding{55} & \ding{55} \\
downscaled view at inference                      & \ding{55} & --- & --- \\
ground-truth-conditioned teacher                  & \checkmark & \ding{55} & \checkmark \\
\textbf{\ours{} (view $+$ hint teacher)} & \checkmark & \checkmark & \checkmark \\
\bottomrule
\end{tabular}
\end{table}
     % budget  900 w   + Fig 2
% =====================================================================
%  4  METHOD                                       BUDGET: ~550 words
% =====================================================================
\section{Method}
\label{sec:method}

\begin{figure}[!tb]
\centering
\includegraphics[width=\textwidth]{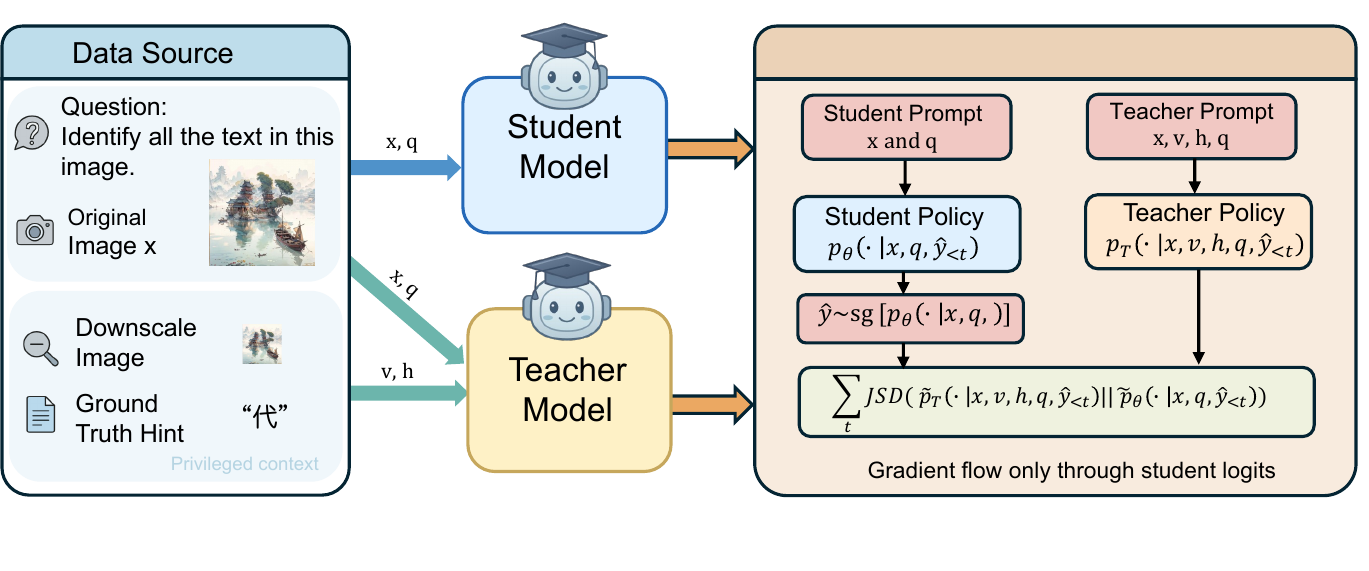}
\caption{\textbf{Overview of \ours{}.} The teacher receives the original image,
the downscaled view $v=g(x)$, the task instruction, and the ground-truth
output as privileged context. Its token distributions supervise the student
along stop-gradient, student-generated trajectories. Gradients flow only
through the student logits, and inference uses the original image alone.}
\label{fig:method}
\end{figure}

Table~\ref{tab:desiderata} suggests supervising the student under a viewing
condition in which Squint Text is easier to read. During training, the teacher
receives privileged context consisting of the downscaled view and the
ground-truth output, while the student receives only the original image.
Inference remains original-only. Figure~\ref{fig:method} illustrates the
training framework.

\subsection{\ours{}}
\label{sec:setup}

Let $x$ be the original image, $v=g(x)$ its downscaled view, $q$ the task
instruction, and $h$ the ground-truth output provided only to the teacher.
The student is $p_\theta(y\mid x,q)$ and the teacher is a frozen copy of the
initial backbone conditioned on $(x,v,q,h)$. At each update, we first sample a
trajectory $\hat y$ from the current student policy and treat the sampled
prefixes as stop-gradient states:
\begin{equation}
\hat y \sim \operatorname{sg}\!\left[p_\theta(\cdot\mid x,q)\right].
\end{equation}
We then minimize token-level Jensen--Shannon divergence between teacher and
student distributions along that student-generated trajectory:
\begin{equation}
\mathcal{L}_{\mathrm{OPSD}}(\theta)=
\mathbb{E}_{x}\,
\mathbb{E}_{\hat y\sim \operatorname{sg}[p_\theta(\cdot\mid x,q)]}
\sum_{t=1}^{T(\hat y)}
\mathrm{JSD}\!\left(
\widetilde p_T(\cdot\mid x,v,q,h,\hat y_{<t})
\,\big\|\,
\widetilde p_\theta(\cdot\mid x,q,\hat y_{<t})
\right),
\label{eq:loss}
\end{equation}
where gradients flow only through the student logits. For each step, the
student's top-$K$ tokens define the explicit support; probability mass outside
this support is aggregated into a single tail bucket for both teacher and
student before computing JSD. We use $K=100$. Rollouts terminate at EOS or the
maximum response length. Full rollout and optimization details are given in
Appendix~\ref{app:training}.

The privileged view and the ground-truth hint play different roles. The view
changes the visual condition under which the teacher reads the image, whereas
the hint supplies privileged label information. Their contributions are
separated empirically in Table~\ref{tab:ablation} and
Appendix~\ref{app:teacher}.

\subsection{Privileged View}
\label{sec:why}

Since $v=g(x)$ is a deterministic, lossy transform of the original image, it
does not introduce new Shannon information conditional on $x$. Its value is
therefore computational \citep{Xu2020A}: the same visual content can become
easier for a fixed model to exploit after high-frequency suppression and scale
change. The
ground-truth hint $h$, by contrast, does provide privileged label information.
Accordingly, we do not assume that the complete teacher distribution
$p_T(\cdot\mid x,v,q,h)$ is exactly realizable from $(x,q)$ alone. Our objective is
empirical: transfer the components of the teacher's reading behavior that the
student can acquire while retaining original-only inference.

% ---- Figure 3: METHOD FRAMEWORK / PIPELINE -----------------------------
        % budget  650 w   + Fig 3, Tab 1
% =====================================================================
%  5  EXPERIMENTS
%  Revised for a shorter, result-driven presentation.
% =====================================================================
\section{Experiments}
\label{sec:experiments}

\subsection{Experimental Setup}
\label{sec:exp-setup}

\paragraph{Datasets and metrics.}
We train on \num{5905} prompts, including \num{3489} synthetic training
images, \num{1213} OCR samples, and \num{1203} VQA samples. Our primary
evaluation is set-matched ANLS on all \num{1076} deployment-derived \bench{}
images. We further
evaluate transfer on Illusory VQA \citep{11147971}, AdvOCR \citep{Xu_2026_CVPR},
and HC-Bench \citep{li-etal-2025-semvink}, and general vision-language capability on AI2D
\citep{10.1007/978-3-319-46493-0_15}, DocVQA \citep{Mathew_2021_WACV}, and OCRBench
\citep{liu2024ocrbench}. We reserve a 1,045-example held-out validation split from the same source mixture as training.

\paragraph{Baselines.}
We compare against the pretrained backbone (\base{}), appending a downscaled
view at inference (\base{} $+1/8$ view, two-view), single-view SFT, and DAPO
\citep{yu2025dapoopensourcellmreinforcement}. For trained baselines,
we focus on methods that use the original image at inference, matching the
deployment constraint of \ours{}. We also report four off-the-shelf MLLMs as
external references: claude-opus-5, gemini-3.1-pro, gpt-6-astra, and
Qwen3-VL-235B-Think.

\paragraph{Model and training.}In the main experiments, all trained methods use Qwen3.5-9B. Unless stated otherwise, all parameters,
including the vision tower, are updated. In the
current experiments, the teacher is a frozen copy of the initial backbone and
receives the privileged downscaled view and ground-truth hint, while the student
receives only the original image. \ours{}, DAPO, and SFT are all trained for
three epochs. Full optimization and rollout
details are in Appendix~\ref{app:training}; the held-out mixture is described in
Appendix~\ref{app:valmix}.

\paragraph{Inference.}
\ours{} receives only the original image at inference. Additional scaled views
are provided only to baselines explicitly marked with them.

\subsection{Main Results}
\label{sec:main-results}

\begin{table}[t]
\centering\small
\caption{\bench{} performance and general vision-language capability.}
\label{tab:main}
\begin{tabular}{@{}lcccc@{}}
\toprule
& \bench{} & \multicolumn{3}{c}{general} \\
\cmidrule(l){3-5}
& & \small AI2D & \small DocVQA & \small OCRBench \\
\midrule
\base{}            & \num{26.50} & \num{86.66} & \num{95.93} & \num{879} \\
\base{} $+1/8$ view (two-view)   & \num{58.96} & \num{52.36} & \num{95.37} & \num{834} \\
% SFT (LoRA, single) & \num{76.70} & \num{85.33} & \num{95.96} & \num{856} \\  % 2026-09-24: superseded by the full-parameter SFT row below; LoRA numbers kept for the record
SFT (full, single) & \num{72.47} & \num{85.52} & \num{64.73} & \num{722} \\
DAPO               & \num{73.55} & \num{82.12} & \num{92.31} & \num{728} \\
\midrule
\ours{}            & \textbf{\num{88.61}} & \num{85.33} & \num{92.98} & \num{833} \\
\bottomrule
\end{tabular}
\end{table}

\begin{table}[t]
\centering\small
\setlength{\tabcolsep}{3.5pt}
\caption{Comparison with the base model and off-the-shelf MLLMs on
\bench{} and three external hidden-text benchmarks. WER is lower better.
The ``only'' conditions replace the original image with the reduced view,
applied before the model preprocessor; further protocol
details are in Appendix~\ref{app:script}.}
\label{tab:mllm}
\begin{tabular}{llcccccc}
\toprule
benchmark & cond.
& \shortstack[c]{\textsc{Squint}\\\textsc{OPSD}}
& \shortstack[c]{Qwen3.5-\\9B base}
& \shortstack[c]{claude-\\opus-5}
& \shortstack[c]{gemini-\\3.1-pro}
& \shortstack[c]{gpt-6-\\astra}
& \shortstack[c]{qwen3-vl-\\235b-think} \\
\midrule
\multirow{3}{*}{\bench{}}
& orig.          & \textbf{\num{88.61}} & \num{26.50} & \num{22.26} & \num{18.15} & \num{8.33}  & \num{24.44} \\
& $1/8$ only     & \textbf{\num{49.64}} & \num{38.22} & \num{22.41} & \num{27.90} & \num{8.13}  & \num{34.37} \\
& $40{\times}40$ only & \textbf{\num{76.12}} & \num{68.60} & \num{11.07} & \num{44.91} & \num{6.26}  & \num{66.88} \\
\midrule
IllVQA \emph{Illusion} & WER$\downarrow$ & \textbf{\num{67.00}} & \num{178.10} & \num{166.50} & \num{190.13} & \num{119.60} & \num{177.97} \\
IllVQA \emph{Filtered} & WER$\downarrow$ & \textbf{\num{64.87}} & \num{76.80}  & \num{83.53}  & \num{70.00}  & \num{65.27}  & \num{91.33} \\
\midrule
\multirow{3}{*}{AdvOCR$^{\dagger}$}
& orig.          & \textbf{\num{66.7}} & \num{3.7}  & \num{3.7}  & \num{3.7}  & \num{11.1} & \num{3.7} \\
& $1/8$ only     & \textbf{\num{70.4}} & \num{44.4} & \num{33.3} & \num{33.3} & \num{0.0}  & \num{51.9} \\
& $40{\times}40$ only & \textbf{\num{70.4}} & \num{51.9} & \num{22.2} & \num{44.4} & \num{44.4} & \num{55.6} \\
\midrule
\multirow{3}{*}{HC-Bench}
& orig.          & \textbf{\num{44.6}} & \num{1.8}  & \num{5.4}  & \num{5.4}  & \num{1.79} & \num{1.8} \\
& $1/8$ only     & \textbf{\num{46.4}} & \num{26.8} & \num{19.6} & \num{33.9} & \num{12.5} & \num{28.6} \\
& $40{\times}40$ only & \num{19.6} & \num{19.6} & \num{8.9} & \num{19.6} & \num{12.5} & \textbf{\num{25.0}} \\
\bottomrule
\end{tabular}
\end{table}

\paragraph{Target performance.}
\ours{} reaches \num{88.61} ANLS on \bench{}, compared with \num{26.50} for
\base{} and \num{58.96} when the corrective view is appended only at inference
(Table~\ref{tab:main}). It also exceeds single-view SFT (\num{72.47}) and
DAPO (\num{73.55}). SFT improves the target metric only partway and forgets
heavily: its DocVQA falls from \num{95.93} to \num{64.73} and OCRBench from
\num{879} to \num{722}.
SFT warm starts likewise underperform direct OPSD, suggesting that first fitting a hard one-hot target 
is less effective than preserving distributional supervision from the outset.
We also ran a series of teacher-view ablations that vary only $v$ in the
fixed teacher input $(x,v,q,h)$; none of the alternative views or their
aggregations improves on the default (Appendix~\ref{app:teacher}).

\textcolor{blue}{On the full \num{1076}-image \bench{}, \ours{} achieves \num{84.09} ANLS on Han-character transcription and \num{94.20} on Latin-letter transcription, outperforming both \base{} and SFT on each script. Appendix~\ref{app:script-transcription} reports the character-projection protocol and full results.}

\textcolor{blue}{Generalization results on additional models are reported in Appendix~\ref{app:model-generalization}.}\paragraph{External transfer.}
The capability transfers to external hidden-text benchmarks
(Table~\ref{tab:mllm}). On original inputs, \ours{} reaches \num{66.7} on
AdvOCR and \num{44.6} on HC-Bench, compared with \num{3.7} and \num{1.8} for
\base{}. On Illusory VQA, it reduces WER from \num{178.10} to \num{67.00},
even outperforming \base{} given the filtered view (\num{76.80}). Increasing
model scale alone also does not remove the failure: Qwen3-VL-235B
\citep{bai2025qwen3vltechnicalreport} reaches only \num{24.44} on the original
\bench{} images, and gpt-6-astra only \num{8.33}.
\textcolor{blue}{Qwen3.5-9B's higher Han-character but nearly identical Latin-letter scores relative to Claude suggest that script composition may contribute to their ranking (Appendix~\ref{app:script-transcription}, Table~\ref{tab:cross-model-script}).}

\paragraph{General-capability trade-off.}
The hidden-text gain is accompanied by task-dependent degradation on the
general benchmarks: AI2D changes from \num{86.66} to \num{85.33}, DocVQA
from \num{95.93} to \num{92.98}, and OCRBench from \num{879} to \num{833}.
Thus, the adaptation retains substantial general vision-language capability,
but introduces a measurable specialization trade-off.

\subsection{Ablations}
\label{sec:ablations}

\begin{table}[t]
\centering\small
\caption{Ablations on \bench{}. We vary teacher input, \textcolor{blue}{trajectory source and supervision}, and
trainable modules.}
\label{tab:ablation}
\begin{tabular}{llc}
\toprule
& configuration & \bench{} \\
\midrule
\multicolumn{3}{@{}l}{\emph{teacher input}}\\
& $x$ \hfill (neither)                              & \num{27.96} \\
& $x+h$ \hfill (hint only)                          & \num{41.79} \\
& $x+v$ \hfill (privileged view only)               & \num{54.68} \\
& $x+v+h$ \hfill (default)                          & \textbf{\num{88.61}} \\
& $x+v_{\rm mis}+h$ \hfill (mismatched view) & \num{78.60} \\
& $x+v+h_{\rm wrong}$ \hfill (wrong hint)           & \num{64.00} \\
\addlinespace
\multicolumn{3}{@{}l}{\textcolor{blue}{\emph{trajectory and supervision}}}\\
& \textcolor{blue}{ground truth, hard (SFT)}          & \textcolor{blue}{\num{72.47}} \\
& \textcolor{blue}{teacher offline, hard (SFT-distil)} & \textcolor{blue}{\num{68.31}} \\
& \textcolor{blue}{teacher offline, soft}            & \textcolor{blue}{\num{77.57}} \\
& \textcolor{blue}{student online, hard}             & \textcolor{blue}{\num{72.67}} \\
& \textcolor{blue}{student online, soft (\ours{})}    & \textcolor{blue}{\textbf{\num{88.61}}} \\
\addlinespace
\multicolumn{3}{@{}l}{\emph{trainable modules (\ours{})}}\\
& LM only, vision tower frozen                       & \num{29.25} \\
& vision tower only, LM frozen                       & \num{71.12} \\
& all parameters                                     & \textbf{\num{88.61}} \\
\bottomrule
\end{tabular}
\end{table}

\paragraph{Privileged teacher context.}
Neither signal alone gets far: hint-only teacher input yields \num{41.79},
view-only yields \num{54.68}, and $x$-only yields \num{27.96}. Combining the
view and hint reaches \num{88.61}, substantially above either signal alone.
Independently, swapping the view for another sample's downscaled image
reduces the paired score from \textcolor{blue}{\num{88.61}} to \num{78.60}, and corrupting the
hint with a wrong transcription drops it to \num{64.00}. The two privileged signals are
therefore complementary, and each must be aligned with the current sample:
the view contributes sample-specific visual context rather than merely extra
low-resolution tokens, while the hint supplies label information beyond what
the view alone provides.

\textcolor{blue}{Teacher diagnostics show lower gold-prefix transcription-token accuracy for the view-only teacher; at selected student empty-answer decisions, the hint-only teacher often supports the impending empty answer, while the full teacher more often favors the correct non-empty answer. These differences help explain why both teacher-input ablations underperform full \ours{} (Appendix~\ref{app:teacher-prefix}).}

\paragraph{\textcolor{blue}{Trajectory and supervision.}}
\textcolor{blue}{To examine the contribution of on-policy distillation, we compare
student-generated online trajectories with cached teacher-generated offline
trajectories under the same teacher input: the original image, a $1/8$-scale
view, and the ground-truth hint.
Online trajectories improve ANLS from \num{77.57} to \num{88.61} with soft
supervision and from \num{68.31} to \num{72.67} with hard supervision.
Soft supervision also outperforms hard supervision under both trajectory
settings, supporting the effectiveness of distributional distillation in our
setup. Section~\ref{sec:runnerup} further examines the teacher distribution.}

\paragraph{Where adaptation occurs.}
Updating only the language side reaches \num{29.25}, whereas updating only the
vision tower already reaches \num{71.12}. Joint adaptation further improves to
\num{88.61}. Thus, visual adaptation accounts for a large part of the gain,
while the strongest read-out requires both sides to adapt. The same dependence
appears under supervised fine-tuning (Appendix~\ref{app:sft-modules}), supporting
the representation change analyzed in \S\ref{sec:why-works}.

\paragraph{Teacher and training variants.}
Entropy-based teacher selection, EMA teacher updates, and SFT warm starts
likewise underperform the fixed-teacher recipe
(Appendix~\ref{app:teacher}).
   % budget  800 w   + Tab 2, Tab 3, Tab 4
% =====================================================================
%  6  WHAT DOES SQUINT-OPSD LEARN?
%  Revised for a shorter, result-driven presentation.
% =====================================================================
\section{Analysis}
\label{sec:why-works}

We analyze three changes induced by training: glyph representation,
spatial selection, and candidate-distribution structure. The probe and
candidate-distribution analyses reuse the protocols established in
\S\ref{sec:diagnosis}.

\paragraph{Glyph representations become more discriminative.}
Re-running the probes from \S\ref{sec:encoded}, glyph-identity AUC rises
from \num{0.718} to \num{0.975} on Chinese and from \num{0.612} to
\num{0.779} on Latin, while near-ceiling presence detection changes little.
\textcolor{blue}{The module-adaptation controls for \ours{} also favor vision-only over LM-only adaptation (\num{71.12} vs.\ \num{29.25} ANLS; \S\ref{sec:ablations}), supporting an important role for representation changes alongside spatial read-out.}

\begin{figure}[t]
\centering
\includegraphics[width=\textwidth]{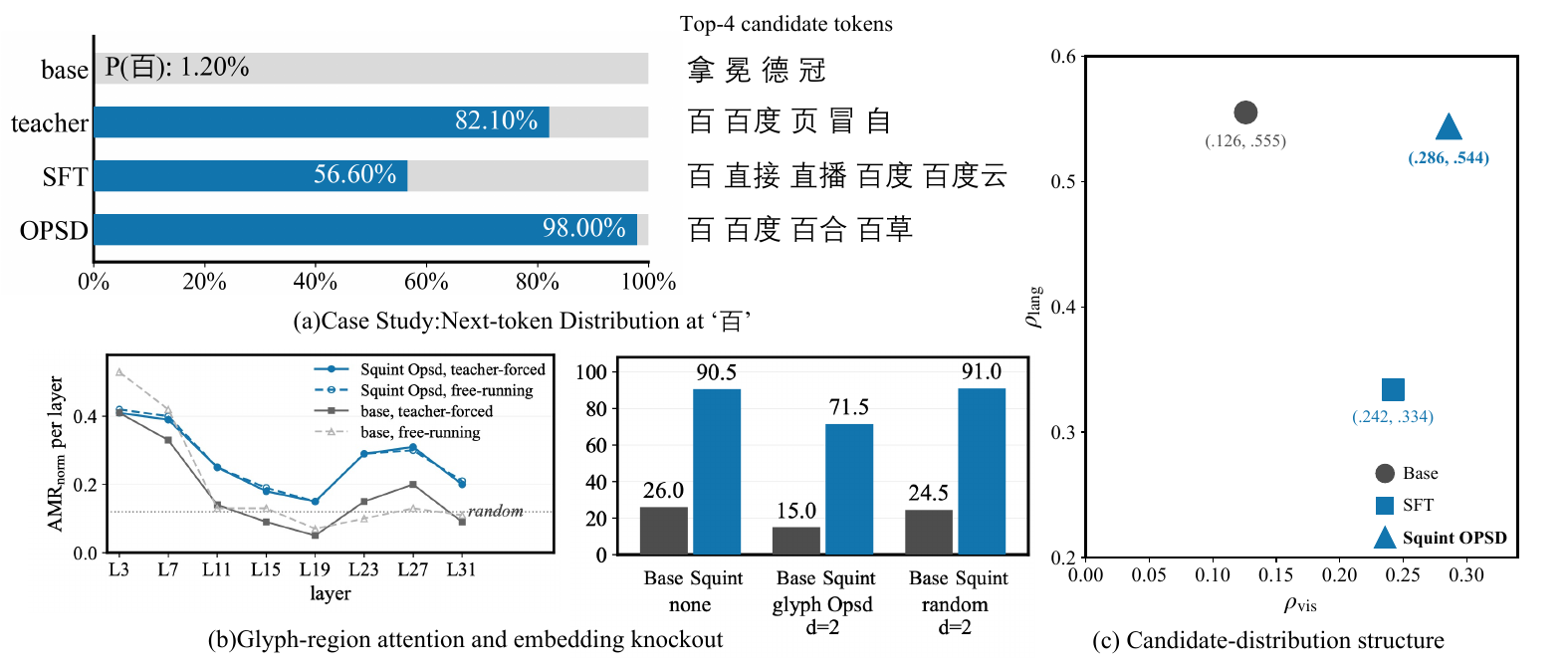}
\caption{\textbf{What changes after training.}
\textbf{(a)} Example next-token distributions for \base{}, SFT, and \ours{}.
\textbf{(b)} Attention concentration on glyph-bearing regions and a causal
visual-token embedding knockout.
\textbf{(c)} Candidate-distribution structure on \num{120} matched glyph
positions.}
\label{fig:analysis}
\end{figure}

\paragraph{Reading becomes spatially grounded.}
\label{sec:amr}
After training, answer-token attention \citep{NIPS2017_3f5ee243} is more
concentrated on glyph-bearing regions, with similar patterns under free
generation and teacher forcing (Figure~\ref{fig:analysis}b). We complement this
correlational evidence with a visual-token embedding intervention. Before
the visual tokens enter the language model, we replace glyph-bearing
embeddings with the mean embedding of non-glyph tokens from the same image,
while keeping token count and positions unchanged. At dilation $d=2$, this
reduces \ours{} accuracy from \num{90.5}\% to \num{71.5}\%, whereas
replacing an equal number of randomly selected background-token embeddings
leaves performance nearly unchanged (\num{90.5}\% to \num{91.0}\%). The
result supports a causal dependence of the learned reading behavior on
glyph-bearing visual regions. Full intervention details are given in
Appendix~\ref{app:knockout}.

\paragraph{The distribution acquires visual structure.}
\label{sec:runnerup}
We compare non-target, high-probability candidates on the same \num{120}
glyph positions valid for all three models. We measure
$\rho_{\mathrm{vis}}$ as the Spearman correlation between candidate
probability and pixel-IoU similarity to the target glyph, and
$\rho_{\mathrm{lang}}$ as the correlation between candidate probability and
its matched text-only probability. Visual organization increases
from $\rho_{\mathrm{vis}}=\num{0.126}$ for \base{} to \num{0.242} for SFT
and \num{0.286} for \ours{}. Meanwhile, \ours{} retains a text-only-prior
correlation close to \base{} ($\rho_{\mathrm{lang}}=\num{0.544}$ vs.\
\num{0.555}), whereas SFT drops to \num{0.334}. Thus, \ours{} more strongly
organizes high-probability alternatives by glyph compatibility while retaining
language-side structure. Definitions and candidate-set sensitivity are in
Appendix~\ref{app:runnerup-metrics}.

% ---- Figure 4: CASE STUDY + ANALYSIS CHARTS ----------------------------

\section{Related Work}
\label{sec:related}

\paragraph{Hidden and squint-visible content.}
HC-Bench studies hidden content with downscaling \citep{li-etal-2025-semvink},
Illusory VQA applies low-pass filtering \citep{11147971}, and
AdvOCR/VACoT learns when to zoom in or out at inference \citep{Xu_2026_CVPR};
training-free alternatives steer a frozen model's search toward small visual
details \citep{Wu_2024_CVPR,zhang2025mllms}.
These works alter the viewing condition or the search process at inference. To
our knowledge,
\ours{} is the first work on squint-visible hidden text to address this
failure through post-training for original-only inference.
\bench{} adds deployment-derived failures and \probeset{} supports controlled
diagnosis; the frequency effect is related to hybrid images
\citep{10.1145/1141911.1141919} and image-scaling attacks \citep{251570},
a principle recently productized by hidden-content fonts \citep{decoyfont},
while text-to-image generators render the coarse-structure channel
increasingly accessible \citep{wu2025qwenimagetechnicalreport}.

\paragraph{Text-rich reading and privileged supervision.}
Prior work improves pixel-text understanding through reading-oriented data
and architectures \citep{Zhang_2024_CVPR}, benchmarks text-rich comprehension
\citep{li2024seedbench2plusbenchmarkingmultimodallarge}, and studies the symbolic-to-pixel modality gap
\citep{sun2026readingthinkingunderstandingbridging,fu-etal-2025-multimodal}. Global--local methods provide
additional visual views \citep{lu2026globallocaldualperceptionmllms}, and native high-resolution
input pipelines enlarge the visual budget available to small text
\citep{DBLP:conf/eccv/GuoXYCNGCLH24,Chen_2024_CVPR}. Related privileged-information
and on-policy distillation methods use signals available only during training
\citep{VAPNIK2009544,hinton2015distillingknowledgeneuralnetwork,agarwal2024onpolicy,zhao2026selfdistilled,yuan2026visionopdlearningfinedetails,li2026visualopsdcrossmodalonpolicyselfdistillation}.
These works largely target fine-grained visual detail retrieval.

\paragraph{Probing visual representations.}
Linear probes test information in intermediate representations
\citep{alain2018understandingintermediatelayersusing,burns2023discovering}, while VLM work localizes object
information to visual tokens \citep{neo2025towards} and identifies
outlier artifact tokens that distort ViT feature maps \citep{darcet2024vision}. We study coarse text whose identity becomes
more accessible at the correct locations and how this changes after adaptation.    % budget  600 w   + Fig 4
\section{Conclusion}

We introduce Squint Text, \bench{}, and \probeset{} to study text embedded in
coarse image structure that humans can recover but MLLMs often miss.
Our diagnostics implicate spatial selection and high-frequency interference,
with visual-token count as a possible additional factor. These findings
motivate \ours{}, which uses a downscaled view and ground-truth transcription
as privileged training context while retaining original-only inference.
\ours{} reaches 88.61 ANLS on \bench{} versus 26.50 for the base model and
transfers to external hidden-text benchmarks with measurable general-capability
trade-offs. After training, glyph identity becomes more discriminative, reading
depends on glyph-bearing regions under our visual-token embedding intervention,
and high-probability alternatives become more visually structured while
retaining language-side structure. The method currently relies on recoverable
visual evidence and fixed privileged supervision.    % budget  300 w   (limitations folded in)

% --- Statements -------------------------------------------------------
%  AI Use Statement (REQUIRED by ICLR 2027), Ethics Statement and
%  Reproducibility Statement (recommended).  All are page-exempt; the
%  Author Guidelines ask for them at the end of the main text, BEFORE the
%  references -- so they go here, right after the body \clearpage.
\clearpage
% Ethics Statement (recommended by ICLR 2027; page-exempt, <= 1 page).
\section*{Ethics Statement}

\bench{} is derived from production content-moderation traffic and contains
domains, contact strings and policy-violating promotional material, so it
raises privacy and dual-use concerns that the fully synthetic \probeset{} suite
does not. The benchmark has been released on Hugging Face,
but because its images embed real websites, access to the production-derived
portion is gated and granted on approved request; we study these patterns
solely to improve the robustness of visual moderation.

% =====================================================================
%  ICLR 2027 STATEMENTS
%    AI Use Statement          -- REQUIRED, page-exempt.
%    Reproducibility Statement -- recommended, page-exempt.
%  Placed at the end of the main text, before the references.
%
%  !! Please confirm the AI Use Statement matches what you actually did.
%     The Hugging Face URL for \bench{} is intentionally omitted during
%     review (a dataset page naming the authors would break anonymity);
%     add it at camera-ready.
% =====================================================================

\section*{AI Use Statement}

We used generative AI tools for code implementation, translation of the Chinese
prompt templates, and language editing of the manuscript. All AI-assisted output was reviewed and verified by the authors against
the raw evaluation logs and the original sources, and we take responsibility
for the final content of this paper.

\section*{Reproducibility Statement}

Code, the evaluation protocol and the synthetic \probeset{} suite are released
at \url{https://anonymous.4open.science/r/Squint_Text-3456}; \bench{} has been released on Hugging Face, with the
production-derived portion gated behind an approved request. Training, optimisation, prompt and preprocessing
details are given in Appendices.

% References start on a fresh page (body ends with blank space before it)
\clearpage
\bibliography{refs}

\begin{thebibliography}{60}
\providecommand{\natexlab}[1]{#1}
\providecommand{\url}[1]{\texttt{#1}}
\expandafter\ifx\csname urlstyle\endcsname\relax
  \providecommand{\doi}[1]{doi: #1}\else
  \providecommand{\doi}{doi: \begingroup \urlstyle{rm}\Url}\fi

\bibitem[Agarwal et~al.(2024)Agarwal, Vieillard, Zhou, Stanczyk, Garea, Geist,
  and Bachem]{agarwal2024onpolicy}
Rishabh Agarwal, Nino Vieillard, Yongchao Zhou, Piotr Stanczyk, Sabela~Ramos
  Garea, Matthieu Geist, and Olivier Bachem.
\newblock On-policy distillation of language models: Learning from
  self-generated mistakes.
\newblock In \emph{The Twelfth International Conference on Learning
  Representations}, 2024.
\newblock URL \url{https://openreview.net/forum?id=3zKtaqxLhW}.

\bibitem[Alain \& Bengio(2018)Alain and
  Bengio]{alain2018understandingintermediatelayersusing}
Guillaume Alain and Yoshua Bengio.
\newblock Understanding intermediate layers using linear classifier probes,
  2018.
\newblock URL \url{https://arxiv.org/abs/1610.01644}.

\bibitem[Bai et~al.(2025)]{bai2025qwen3vltechnicalreport}
Shuai Bai et~al.
\newblock Qwen3-vl technical report, 2025.
\newblock URL \url{https://arxiv.org/abs/2511.21631}.

\bibitem[Biten et~al.(2019)Biten, Tito, Mafla, Gomez, Rusinol, Valveny,
  Jawahar, and Karatzas]{Biten_2019_ICCV}
Ali~Furkan Biten, Ruben Tito, Andres Mafla, Lluis Gomez, Marcal Rusinol, Ernest
  Valveny, C.V. Jawahar, and Dimosthenis Karatzas.
\newblock Scene text visual question answering.
\newblock In \emph{Proceedings of the IEEE/CVF International Conference on
  Computer Vision (ICCV)}, October 2019.

\bibitem[Burns et~al.(2023)Burns, Ye, Klein, and
  Steinhardt]{burns2023discovering}
Collin Burns, Haotian Ye, Dan Klein, and Jacob Steinhardt.
\newblock Discovering latent knowledge in language models without supervision.
\newblock In \emph{The Eleventh International Conference on Learning
  Representations}, 2023.
\newblock URL \url{https://openreview.net/forum?id=ETKGuby0hcs}.

\bibitem[Chen et~al.(2024)Chen, Wu, Wang, Su, Chen, Xing, Zhong, Zhang, Zhu,
  Lu, Li, Luo, Lu, Qiao, and Dai]{Chen_2024_CVPR}
Zhe Chen, Jiannan Wu, Wenhai Wang, Weijie Su, Guo Chen, Sen Xing, Muyan Zhong,
  Qinglong Zhang, Xizhou Zhu, Lewei Lu, Bin Li, Ping Luo, Tong Lu, Yu~Qiao, and
  Jifeng Dai.
\newblock Internvl: Scaling up vision foundation models and aligning for
  generic visual-linguistic tasks.
\newblock In \emph{Proceedings of the IEEE/CVF Conference on Computer Vision
  and Pattern Recognition (CVPR)}, pp.\  24185--24198, June 2024.

\bibitem[Darcet et~al.(2024)Darcet, Oquab, Mairal, and
  Bojanowski]{darcet2024vision}
Timoth{\'e}e Darcet, Maxime Oquab, Julien Mairal, and Piotr Bojanowski.
\newblock Vision transformers need registers.
\newblock In \emph{The Twelfth International Conference on Learning
  Representations}, 2024.
\newblock URL \url{https://openreview.net/forum?id=2dnO3LLiJ1}.

\bibitem[Fu et~al.(2025)Fu, Guan, Wang, Guo, Duan, Sun, Chen, Jiang, Ma, Zhou,
  and Luo]{fu-etal-2025-multimodal}
Pei Fu, Tongkun Guan, Zining Wang, Zhentao Guo, Chen Duan, Hao Sun, Boming
  Chen, Qianyi Jiang, Jiayao Ma, Kai Zhou, and Junfeng Luo.
\newblock Multimodal large language models for text-rich image understanding: A
  comprehensive review.
\newblock In Wanxiang Che, Joyce Nabende, Ekaterina Shutova, and Mohammad~Taher
  Pilehvar (eds.), \emph{Findings of the Association for Computational
  Linguistics: ACL 2025}, pp.\  19941--19958, Vienna, Austria, July 2025.
  Association for Computational Linguistics.
\newblock ISBN 979-8-89176-256-5.
\newblock \doi{10.18653/v1/2025.findings-acl.1023}.
\newblock URL \url{https://aclanthology.org/2025.findings-acl.1023/}.

\bibitem[Guo et~al.(2024{\natexlab{a}})Guo, Utkarsh, Ding, Ondracek, Zhao,
  Freeman, Vishwamitra, and Hu]{299571}
Keyan Guo, Ayush Utkarsh, Wenbo Ding, Isabelle Ondracek, Ziming Zhao, Guo
  Freeman, Nishant Vishwamitra, and Hongxin Hu.
\newblock Moderating illicit online image promotion for unsafe user generated
  content games using large {Vision-Language} models.
\newblock In \emph{33rd USENIX Security Symposium (USENIX Security 24)}, pp.\
  5787--5804, Philadelphia, PA, August 2024{\natexlab{a}}. USENIX Association.
\newblock ISBN 978-1-939133-44-1.
\newblock URL
  \url{https://www.usenix.org/conference/usenixsecurity24/presentation/guo-keyan}.

\bibitem[Guo et~al.(2024{\natexlab{b}})Guo, Xu, Yao, Cui, Ni, Ge, Chua, Liu,
  and Huang]{DBLP:conf/eccv/GuoXYCNGCLH24}
Zonghao Guo, Ruyi Xu, Yuan Yao, Junbo Cui, Zanlin Ni, Chunjiang Ge, Tat{-}Seng
  Chua, Zhiyuan Liu, and Gao Huang.
\newblock Llava-uhd: An {LMM} perceiving any aspect ratio and high-resolution
  images.
\newblock In Ales Leonardis, Elisa Ricci, Stefan Roth, Olga Russakovsky,
  Torsten Sattler, and G{\"{u}}l Varol (eds.), \emph{Computer Vision - {ECCV}
  2024 - 18th European Conference, Milan, Italy, September 29-October 4, 2024,
  Proceedings, Part {LXXXIII}}, volume 15141 of \emph{Lecture Notes in Computer
  Science}, pp.\  390--406. Springer, 2024{\natexlab{b}}.
\newblock \doi{10.1007/978-3-031-73010-8\_23}.
\newblock URL \url{https://doi.org/10.1007/978-3-031-73010-8\_23}.

\bibitem[Hinton et~al.(2015)Hinton, Vinyals, and
  Dean]{hinton2015distillingknowledgeneuralnetwork}
Geoffrey Hinton, Oriol Vinyals, and Jeff Dean.
\newblock Distilling the knowledge in a neural network, 2015.
\newblock URL \url{https://arxiv.org/abs/1503.02531}.

\bibitem[Jaume et~al.(2019)Jaume, Ekenel, and Thiran]{8892998}
Guillaume Jaume, Hazim~Kemal Ekenel, and Jean-Philippe Thiran.
\newblock Funsd: A dataset for form understanding in noisy scanned documents.
\newblock In \emph{2019 International Conference on Document Analysis and
  Recognition Workshops (ICDARW)}, volume~2, pp.\  1--6, 2019.
\newblock \doi{10.1109/ICDARW.2019.10029}.

\bibitem[Karatzas et~al.(2013)Karatzas, Shafait, Uchida, Iwamura, i~Bigorda,
  Robles~Mestre, Mas, Fernandez~Mota, Almazan~Almazan, and Pere de~las
  Heras]{karatzas2013icdar}
Dimosthenis Karatzas, Faisal Shafait, Seiichi Uchida, Masakazu Iwamura,
  Lluis~Gomez i~Bigorda, Sergi Robles~Mestre, Joan Mas, David Fernandez~Mota,
  Jon Almazan~Almazan, and Lluis Pere de~las Heras.
\newblock Icdar 2013 robust reading competition.
\newblock In \emph{2013 12th International Conference on Document Analysis and
  Recognition (ICDAR)}, pp.\  1484--1493, 2013.

\bibitem[Kembhavi et~al.(2016)Kembhavi, Salvato, Kolve, Seo, Hajishirzi, and
  Farhadi]{10.1007/978-3-319-46493-0_15}
Aniruddha Kembhavi, Mike Salvato, Eric Kolve, Minjoon Seo, Hannaneh Hajishirzi,
  and Ali Farhadi.
\newblock A diagram is worth a dozen images.
\newblock In Bastian Leibe, Jiri Matas, Nicu Sebe, and Max Welling (eds.),
  \emph{Computer Vision -- ECCV 2016}, pp.\  235--251, Cham, 2016. Springer
  International Publishing.
\newblock ISBN 978-3-319-46493-0.

\bibitem[Kiela et~al.(2020)Kiela, Firooz, Mohan, Goswami, Singh, Ringshia, and
  Testuggine]{NEURIPS2020_1b84c4ce}
Douwe Kiela, Hamed Firooz, Aravind Mohan, Vedanuj Goswami, Amanpreet Singh,
  Pratik Ringshia, and Davide Testuggine.
\newblock The hateful memes challenge: Detecting hate speech in multimodal
  memes.
\newblock In H.~Larochelle, M.~Ranzato, R.~Hadsell, M.F. Balcan, and H.~Lin
  (eds.), \emph{Advances in Neural Information Processing Systems}, volume~33,
  pp.\  2611--2624. Curran Associates, Inc., 2020.
\newblock URL
  \url{https://proceedings.neurips.cc/paper_files/paper/2020/file/1b84c4cee2b8b3d823b30e2d604b1878-Paper.pdf}.

\bibitem[Lanczos(1964)]{doi:10.1137/0701007}
C.~Lanczos.
\newblock Evaluation of noisy data.
\newblock \emph{Journal of the Society for Industrial and Applied Mathematics
  Series B Numerical Analysis}, 1\penalty0 (1):\penalty0 76--85, 1964.
\newblock \doi{10.1137/0701007}.
\newblock URL \url{https://doi.org/10.1137/0701007}.

\bibitem[Li et~al.(2024)Li, Ge, Chen, Ge, Zhang, and
  Shan]{li2024seedbench2plusbenchmarkingmultimodallarge}
Bohao Li, Yuying Ge, Yi~Chen, Yixiao Ge, Ruimao Zhang, and Ying Shan.
\newblock Seed-bench-2-plus: Benchmarking multimodal large language models with
  text-rich visual comprehension, 2024.
\newblock URL \url{https://arxiv.org/abs/2404.16790}.

\bibitem[Li et~al.(2026)Li, Gao, Zhang, Huang, Li, Yang, Xu, and
  Liu]{li2026visualopsdcrossmodalonpolicyselfdistillation}
Pengyu Li, Zhitao Gao, Lingling Zhang, Muye Huang, Yuanming Li, Zesheng Yang,
  Fangzhi Xu, and Jun Liu.
\newblock Visual-opsd: Cross-modal on-policy self-distillation for efficient
  unified multimodal reasoning, 2026.
\newblock URL \url{https://arxiv.org/abs/2606.18974}.

\bibitem[Li et~al.(2025)Li, Cai, and Wang]{li-etal-2025-semvink}
Sifan Li, Yujun Cai, and Yiwei Wang.
\newblock {S}em{V}ink: Advancing {VLM}s' semantic understanding of optical
  illusions via visual global thinking.
\newblock In Christos Christodoulopoulos, Tanmoy Chakraborty, Carolyn Rose, and
  Violet Peng (eds.), \emph{Proceedings of the 2025 Conference on Empirical
  Methods in Natural Language Processing}, pp.\  27167--27177, Suzhou, China,
  November 2025. Association for Computational Linguistics.
\newblock ISBN 979-8-89176-332-6.
\newblock \doi{10.18653/v1/2025.emnlp-main.1381}.
\newblock URL \url{https://aclanthology.org/2025.emnlp-main.1381/}.

\bibitem[Liu et~al.(2011)Liu, Yin, Wang, and Wang]{6065272}
Cheng-Lin Liu, Fei Yin, Da-Han Wang, and Qiu-Feng Wang.
\newblock Casia online and offline chinese handwriting databases.
\newblock In \emph{2011 International Conference on Document Analysis and
  Recognition}, pp.\  37--41, 2011.
\newblock \doi{10.1109/ICDAR.2011.17}.

\bibitem[Liu et~al.(2024)Liu, Li, Huang, Yang, Yu, Li, Yin, Liu, Jin, and
  Bai]{liu2024ocrbench}
Yuliang Liu, Zhang Li, Mingxin Huang, Biao Yang, Wenwen Yu, Chunyuan Li,
  Xu-Cheng Yin, Cheng-Lin Liu, Lianwen Jin, and Xiang Bai.
\newblock Ocrbench: On the hidden mystery of ocr in large multimodal models.
\newblock \emph{Science China Information Sciences}, 67\penalty0 (12):\penalty0
  220102, 2024.
\newblock \doi{10.1007/s11432-024-4235-6}.

\bibitem[Loshchilov \& Hutter(2019)Loshchilov and
  Hutter]{loshchilov2018decoupled}
Ilya Loshchilov and Frank Hutter.
\newblock Decoupled weight decay regularization.
\newblock In \emph{International Conference on Learning Representations}, 2019.
\newblock URL \url{https://openreview.net/forum?id=Bkg6RiCqY7}.

\bibitem[Lu et~al.(2026)Lu, Song, Wu, Li, Liang, Yang, Chen, Xie, Lu, Zhao,
  Sun, and Wei]{lu2026globallocaldualperceptionmllms}
Junxin Lu, Tengfei Song, Zhanglin Wu, Pengfei Li, Xiaowei Liang, Hui Yang, Kun
  Chen, Ning Xie, Yunfei Lu, Jing Zhao, Shiliang Sun, and Daimeng Wei.
\newblock Global-local dual perception for mllms in high-resolution text-rich
  image translation, 2026.
\newblock URL \url{https://arxiv.org/abs/2602.21956}.

\bibitem[Mathew et~al.(2021)Mathew, Karatzas, and Jawahar]{Mathew_2021_WACV}
Minesh Mathew, Dimosthenis Karatzas, and C.V. Jawahar.
\newblock Docvqa: A dataset for vqa on document images.
\newblock In \emph{Proceedings of the IEEE/CVF Winter Conference on
  Applications of Computer Vision (WACV)}, pp.\  2200--2209, January 2021.

\bibitem[{Meta}(2026{\natexlab{a}})]{metacser}
{Meta}.
\newblock Community standards enforcement report.
\newblock
  \url{https://transparency.meta.com/reports/community-standards-enforcement/},
  2026{\natexlab{a}}.

\bibitem[{Meta}(2026{\natexlab{b}})]{metahateful}
{Meta}.
\newblock Community standards: Hateful conduct.
\newblock
  \url{https://transparency.meta.com/policies/community-standards/hate-speech/},
  2026{\natexlab{b}}.

\bibitem[{Meta}(2026{\natexlab{c}})]{metarestricted}
{Meta}.
\newblock Community standards: Restricted goods and services.
\newblock
  \url{https://transparency.meta.com/policies/community-standards/regulated-goods/},
  2026{\natexlab{c}}.

\bibitem[Mishra et~al.(2019)Mishra, Shekhar, Singh, and
  Chakraborty]{mishraICDAR19}
Anand Mishra, Shashank Shekhar, Ajeet~Kumar Singh, and Anirban Chakraborty.
\newblock Ocr-vqa: Visual question answering by reading text in images.
\newblock In \emph{ICDAR}, 2019.

\bibitem[{Mixfont}(2026)]{decoyfont}
{Mixfont}.
\newblock Decoy font.
\newblock \url{https://mixfont.com}, 2026.
\newblock TTF font released July 15, 2026.

\bibitem[{Monster Labs}(2023)]{qrmonster}
{Monster Labs}.
\newblock {QR Code Monster v2}: {ControlNet} for {Stable Diffusion} 1.5
  ({control\_v1p\_sd15\_qrcode\_monster}).
\newblock
  \url{https://huggingface.co/monster-labs/control_v1p_sd15_qrcode_monster},
  2023.

\bibitem[Neo et~al.(2025)Neo, Ong, Torr, Geva, Krueger, and
  Barez]{neo2025towards}
Clement Neo, Luke Ong, Philip Torr, Mor Geva, David Krueger, and Fazl Barez.
\newblock Towards interpreting visual information processing in vision-language
  models.
\newblock In \emph{The Thirteenth International Conference on Learning
  Representations}, 2025.
\newblock URL \url{https://openreview.net/forum?id=chanJGoa7f}.

\bibitem[Oliva et~al.(2006)Oliva, Torralba, and
  Schyns]{10.1145/1141911.1141919}
Aude Oliva, Antonio Torralba, and Philippe~G. Schyns.
\newblock Hybrid images.
\newblock \emph{ACM Trans. Graph.}, 25\penalty0 (3):\penalty0 527–532, July
  2006.
\newblock ISSN 0730-0301.
\newblock \doi{10.1145/1141911.1141919}.
\newblock URL \url{https://doi.org/10.1145/1141911.1141919}.

\bibitem[Pizer et~al.(1987)Pizer, Amburn, Austin, Cromartie, Geselowitz, Greer,
  {ter Haar Romeny}, Zimmerman, and Zuiderveld]{PIZER1987355}
Stephen~M. Pizer, E.~Philip Amburn, John~D. Austin, Robert Cromartie, Ari
  Geselowitz, Trey Greer, Bart {ter Haar Romeny}, John~B. Zimmerman, and Karel
  Zuiderveld.
\newblock Adaptive histogram equalization and its variations.
\newblock \emph{Computer Vision, Graphics, and Image Processing}, 39\penalty0
  (3):\penalty0 355--368, 1987.
\newblock ISSN 0734-189X.
\newblock \doi{https://doi.org/10.1016/S0734-189X(87)80186-X}.
\newblock URL
  \url{https://www.sciencedirect.com/science/article/pii/S0734189X8780186X}.

\bibitem[Quiring et~al.(2020)Quiring, Klein, Arp, Johns, and Rieck]{251570}
Erwin Quiring, David Klein, Daniel Arp, Martin Johns, and Konrad Rieck.
\newblock Adversarial preprocessing: Understanding and preventing
  {Image-Scaling} attacks in machine learning.
\newblock In \emph{29th USENIX Security Symposium (USENIX Security 20)}, pp.\
  1363--1380. USENIX Association, August 2020.
\newblock ISBN 978-1-939133-17-5.
\newblock URL
  \url{https://www.usenix.org/conference/usenixsecurity20/presentation/quiring}.

\bibitem[{Qwen Team}(2026)]{qwen35}
{Qwen Team}.
\newblock {Qwen3.5-9B}.
\newblock \url{https://huggingface.co/Qwen/Qwen3.5-9B}, 2026.

\bibitem[Rombach et~al.(2022)Rombach, Blattmann, Lorenz, Esser, and
  Ommer]{Rombach_2022_CVPR}
Robin Rombach, Andreas Blattmann, Dominik Lorenz, Patrick Esser, and Bj\"orn
  Ommer.
\newblock High-resolution image synthesis with latent diffusion models.
\newblock In \emph{Proceedings of the IEEE/CVF Conference on Computer Vision
  and Pattern Recognition (CVPR)}, pp.\  10684--10695, June 2022.

\bibitem[Rostamkhani et~al.(2025)Rostamkhani, Ansari, Sabzevari, Rahmani, and
  Eetemadi]{11147971}
Mohammadmostafa Rostamkhani, Baktash Ansari, Hoorieh Sabzevari, Farzan Rahmani,
  and Sauleh Eetemadi.
\newblock Illusory vqa: Benchmarking and enhancing multimodal models on visual
  illusions.
\newblock In \emph{2025 IEEE/CVF Conference on Computer Vision and Pattern
  Recognition Workshops (CVPRW)}, pp.\  2986--2995, 2025.
\newblock \doi{10.1109/CVPRW67362.2025.00282}.

\bibitem[Shao et~al.(2024)Shao, Wang, Zhu, Xu, Song, Bi, Zhang, Zhang, Li, Wu,
  and Guo]{shao2024deepseekmathpushinglimitsmathematical}
Zhihong Shao, Peiyi Wang, Qihao Zhu, Runxin Xu, Junxiao Song, Xiao Bi, Haowei
  Zhang, Mingchuan Zhang, Y.~K. Li, Y.~Wu, and Daya Guo.
\newblock Deepseekmath: Pushing the limits of mathematical reasoning in open
  language models, 2024.
\newblock URL \url{https://arxiv.org/abs/2402.03300}.

\bibitem[Shi et~al.(2017)Shi, Yao, Liao, Yang, Xu, Cui, Belongie, Lu, and
  Bai]{8270164}
Baoguang Shi, Cong Yao, Minghui Liao, Mingkun Yang, Pei Xu, Linyan Cui, Serge
  Belongie, Shijian Lu, and Xiang Bai.
\newblock Icdar2017 competition on reading chinese text in the wild (rctw-17).
\newblock In \emph{2017 14th IAPR International Conference on Document Analysis
  and Recognition (ICDAR)}, volume~01, pp.\  1429--1434, 2017.
\newblock \doi{10.1109/ICDAR.2017.233}.

\bibitem[Sidorov et~al.(2020)Sidorov, Hu, Rohrbach, and
  Singh]{10.1007/978-3-030-58536-5_44}
Oleksii Sidorov, Ronghang Hu, Marcus Rohrbach, and Amanpreet Singh.
\newblock Textcaps: A dataset for image captioning with reading comprehension.
\newblock In Andrea Vedaldi, Horst Bischof, Thomas Brox, and Jan-Michael Frahm
  (eds.), \emph{Computer Vision -- ECCV 2020}, pp.\  742--758, Cham, 2020.
  Springer International Publishing.
\newblock ISBN 978-3-030-58536-5.

\bibitem[Singh et~al.(2026)]{singh2026openaigpt5card}
Aaditya Singh et~al.
\newblock Openai gpt-5 system card, 2026.
\newblock URL \url{https://arxiv.org/abs/2601.03267}.

\bibitem[Sun et~al.(2026)Sun, Yuan, Liu, Zhao, Zhang, Dredze, and
  Bai]{sun2026readingthinkingunderstandingbridging}
Kaiser Sun, Xiaochuang Yuan, Hongjun Liu, Chen Zhao, Cheng Zhang, Mark Dredze,
  and Fan Bai.
\newblock Reading, not thinking: Understanding and bridging the modality gap
  when text becomes pixels in multimodal llms, 2026.
\newblock URL \url{https://arxiv.org/abs/2603.09095}.

\bibitem[Sun et~al.(2023)Sun, Pan, Ge, Li, Duan, Wu, Zhang, Zhou, Qin, Wang,
  Dai, Qiao, Wang, and Li]{DBLP:conf/nips/SunPGLDWZZQWDQW23}
Keqiang Sun, Junting Pan, Yuying Ge, Hao Li, Haodong Duan, Xiaoshi Wu, Renrui
  Zhang, Aojun Zhou, Zipeng Qin, Yi~Wang, Jifeng Dai, Yu~Qiao, Limin Wang, and
  Hongsheng Li.
\newblock Journeydb: {A} benchmark for generative image understanding.
\newblock In Alice Oh, Tristan Naumann, Amir Globerson, Kate Saenko, Moritz
  Hardt, and Sergey Levine (eds.), \emph{Advances in Neural Information
  Processing Systems 36: Annual Conference on Neural Information Processing
  Systems 2023, NeurIPS 2023, New Orleans, LA, USA, December 10 - 16, 2023},
  2023.
\newblock URL
  \url{http://papers.nips.cc/paper\_files/paper/2023/hash/9bc59aff4685e39e1a8175d5303248a1-Abstract-Datasets\_and\_Benchmarks.html}.

\bibitem[{TikTok}(2026)]{tiktokrca}
{TikTok}.
\newblock Community guidelines: Regulated commercial activities and regulated
  goods.
\newblock
  \url{https://www.tiktok.com/community-guidelines/en/regulated-commercial-activities},
  2026.

\bibitem[Tomasi \& Manduchi(1998)Tomasi and Manduchi]{710815}
C.~Tomasi and R.~Manduchi.
\newblock Bilateral filtering for gray and color images.
\newblock In \emph{Sixth International Conference on Computer Vision (IEEE Cat.
  No.98CH36271)}, pp.\  839--846, 1998.
\newblock \doi{10.1109/ICCV.1998.710815}.

\bibitem[Vapnik \& Vashist(2009)Vapnik and Vashist]{VAPNIK2009544}
Vladimir Vapnik and Akshay Vashist.
\newblock A new learning paradigm: Learning using privileged information.
\newblock \emph{Neural Networks}, 22\penalty0 (5):\penalty0 544--557, 2009.
\newblock ISSN 0893-6080.
\newblock \doi{https://doi.org/10.1016/j.neunet.2009.06.042}.
\newblock URL
  \url{https://www.sciencedirect.com/science/article/pii/S0893608009001130}.
\newblock Advances in Neural Networks Research: IJCNN2009.

\bibitem[Vaswani et~al.(2017)Vaswani, Shazeer, Parmar, Uszkoreit, Jones, Gomez,
  Kaiser, and Polosukhin]{NIPS2017_3f5ee243}
Ashish Vaswani, Noam Shazeer, Niki Parmar, Jakob Uszkoreit, Llion Jones,
  Aidan~N Gomez, \L~ukasz Kaiser, and Illia Polosukhin.
\newblock Attention is all you need.
\newblock In I.~Guyon, U.~Von Luxburg, S.~Bengio, H.~Wallach, R.~Fergus,
  S.~Vishwanathan, and R.~Garnett (eds.), \emph{Advances in Neural Information
  Processing Systems}, volume~30. Curran Associates, Inc., 2017.
\newblock URL
  \url{https://proceedings.neurips.cc/paper_files/paper/2017/file/3f5ee243547dee91fbd053c1c4a845aa-Paper.pdf}.

\bibitem[Wu et~al.(2025{\natexlab{a}})]{wu2025qwenimagetechnicalreport}
Chenfei Wu et~al.
\newblock Qwen-image technical report, 2025{\natexlab{a}}.
\newblock URL \url{https://arxiv.org/abs/2508.02324}.

\bibitem[Wu et~al.(2025{\natexlab{b}})Wu, Zhao, Cao, Xu, Jiang, Wang, Li, Hu,
  Qin, and Fu]{wu2025icmassistantinstructiontuningmultimodallarge}
Mengyang Wu, Yuzhi Zhao, Jialun Cao, Mingjie Xu, Zhongming Jiang, Xuehui Wang,
  Qinbin Li, Guangneng Hu, Shengchao Qin, and Chi-Wing Fu.
\newblock Icm-assistant: Instruction-tuning multimodal large language models
  for rule-based explainable image content moderation, 2025{\natexlab{b}}.
\newblock URL \url{https://arxiv.org/abs/2412.18216}.

\bibitem[Wu \& Xie(2024)Wu and Xie]{Wu_2024_CVPR}
Penghao Wu and Saining Xie.
\newblock {V*}: Guided visual search as a core mechanism in multimodal llms.
\newblock In \emph{Proceedings of the IEEE/CVF Conference on Computer Vision
  and Pattern Recognition (CVPR)}, pp.\  13084--13094, June 2024.

\bibitem[{X Corp.}(2026)]{xillegal}
{X Corp.}
\newblock Rules: Illegal or certain regulated goods and services.
\newblock
  \url{https://help.x.com/en/rules-and-policies/regulated-goods-services},
  2026.

\bibitem[Xu et~al.(2020)Xu, Zhao, Song, Stewart, and Ermon]{Xu2020A}
Yilun Xu, Shengjia Zhao, Jiaming Song, Russell Stewart, and Stefano Ermon.
\newblock A theory of usable information under computational constraints.
\newblock In \emph{International Conference on Learning Representations}, 2020.
\newblock URL \url{https://openreview.net/forum?id=r1eBeyHFDH}.

\bibitem[Xu et~al.(2026)Xu, Sun, Du, Li, Lyu, and Yuan]{Xu_2026_CVPR}
Zhengzhuo Xu, Chong Sun, SiNan Du, Chen Li, Jing Lyu, and Chun Yuan.
\newblock Vacot: Rethinking visual data augmentation with vlms.
\newblock In \emph{Proceedings of the IEEE/CVF Conference on Computer Vision
  and Pattern Recognition (CVPR) Findings}, pp.\  9780--9790, June 2026.

\bibitem[Yu et~al.(2025)]{yu2025dapoopensourcellmreinforcement}
Qiying Yu et~al.
\newblock Dapo: An open-source llm reinforcement learning system at scale,
  2025.
\newblock URL \url{https://arxiv.org/abs/2503.14476}.

\bibitem[Yuan et~al.(2026)Yuan, Lou, Yu, Lin, Sun, Han, and
  Lu]{yuan2026visionopdlearningfinedetails}
Qianhao Yuan, Jie Lou, Xing Yu, Hongyu Lin, Le~Sun, Xianpei Han, and Yaojie Lu.
\newblock Vision-opd: Learning to see fine details for multimodal llms via
  on-policy self-distillation, 2026.
\newblock URL \url{https://arxiv.org/abs/2605.18740}.

\bibitem[Zhang et~al.(2025)Zhang, Khayatkhoei, Chhikara, and
  Ilievski]{zhang2025mllms}
Jiarui Zhang, Mahyar Khayatkhoei, Prateek Chhikara, and Filip Ilievski.
\newblock {MLLM}s know where to look: Training-free perception of small visual
  details with multimodal {LLM}s.
\newblock In \emph{The Thirteenth International Conference on Learning
  Representations}, 2025.
\newblock URL \url{https://openreview.net/forum?id=DgaY5mDdmT}.

\bibitem[Zhang et~al.(2023)Zhang, Rao, and Agrawala]{Zhang_2023_ICCV}
Lvmin Zhang, Anyi Rao, and Maneesh Agrawala.
\newblock Adding conditional control to text-to-image diffusion models.
\newblock In \emph{Proceedings of the IEEE/CVF International Conference on
  Computer Vision (ICCV)}, pp.\  3836--3847, October 2023.

\bibitem[Zhang et~al.(2024)Zhang, Zhang, Chen, Zhou, Gu, Chen, and
  Sun]{Zhang_2024_CVPR}
Ruiyi Zhang, Yanzhe Zhang, Jian Chen, Yufan Zhou, Jiuxiang Gu, Changyou Chen,
  and Tong Sun.
\newblock Trins: Towards multimodal language models that can read.
\newblock In \emph{Proceedings of the IEEE/CVF Conference on Computer Vision
  and Pattern Recognition (CVPR)}, pp.\  22584--22594, June 2024.

\bibitem[Zhao et~al.(2026)Zhao, Xie, Liu, Huang, Pang, Chen, and
  Grover]{zhao2026selfdistilled}
Siyan Zhao, Zhihui Xie, Mengchen Liu, Jing Huang, Guan Pang, Feiyu Chen, and
  Aditya Grover.
\newblock Self-distilled reasoner: On-policy self-distillation for large
  language models.
\newblock In \emph{Forty-third International Conference on Machine Learning},
  2026.
\newblock URL \url{https://openreview.net/forum?id=Jpxfof0EaS}.

\bibitem[Zhao et~al.(2023)Zhao, Gu, Varma, Luo, Huang, Xu, Wright, Shojanazeri,
  Ott, Shleifer, Desmaison, Balioglu, Damania, Nguyen, Chauhan, Hao, Mathews,
  and Li]{10.14778/3611540.3611569}
Yanli Zhao, Andrew Gu, Rohan Varma, Liang Luo, Chien-Chin Huang, Min Xu, Less
  Wright, Hamid Shojanazeri, Myle Ott, Sam Shleifer, Alban Desmaison, Can
  Balioglu, Pritam Damania, Bernard Nguyen, Geeta Chauhan, Yuchen Hao, Ajit
  Mathews, and Shen Li.
\newblock Pytorch fsdp: Experiences on scaling fully sharded data parallel.
\newblock \emph{Proc. VLDB Endow.}, 16\penalty0 (12):\penalty0 3848–3860,
  August 2023.
\newblock ISSN 2150-8097.
\newblock \doi{10.14778/3611540.3611569}.
\newblock URL \url{https://doi.org/10.14778/3611540.3611569}.

\end{thebibliography}
\bibliographystyle{iclr2027_conference}

% Appendix starts on its own page after the references
\clearpage
% =====================================================================
%  APPENDIX -- streamlined version
%
%  Organization:
%  A. Diagnosis controls
%  B. Frequency / preprocessing controls
%  C. Teacher and distillation controls
%  D. Recoverability boundary
%  E. Analysis details
%  F. Training / reproducibility
% =====================================================================
\appendix

% =====================================================================
\section{Additional Diagnosis Controls}
\label{app:diagnosis-plots}

\subsection{Per-layer localization}
\begin{figure}[h]
\centering
\includegraphics[width=0.9\textwidth]{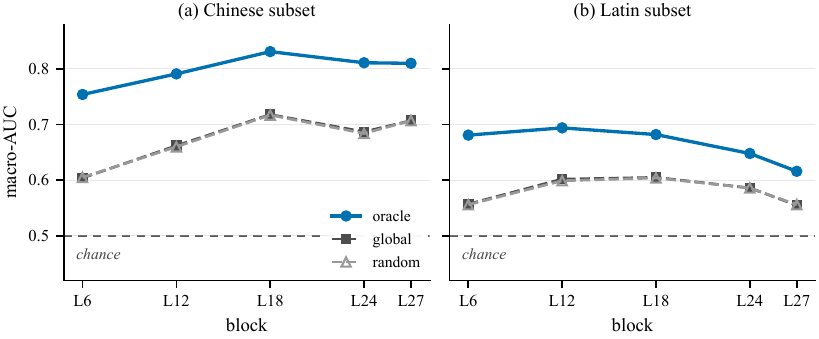}
\caption{\textbf{Glyph identity is more accessible at the correct locations.}
Glyph-identity macro-AUC on \ctrlglyph{} by encoder block under the three
poolings of \S\ref{sec:localization}. \textsc{random} uses the same number of
tokens as \textsc{oracle} but remains close to \textsc{global} across depth and
script.}
\label{fig:localization}
\end{figure}

Figure~\ref{fig:localization} expands Figure~\ref{fig:diagnosis}a across encoder
depth and script. The ordering is stable: restricting the probe to
glyph-bearing tokens consistently improves glyph-identity separability, while
\textsc{random} closely tracks \textsc{global}. Both scripts peak in the middle
of the encoder, but the localization gap persists across depth.

\subsection{Patch-merger control}
\label{app:layers}
The probes in \S\ref{sec:encoded} read pre-merger features, whereas the language
model consumes features after a $2{\times}2$ patch merger. Table~\ref{tab:merger}
shows that global glyph separability changes by at most \num{0.018} AUC across
the merger. The localization gap therefore cannot be explained by information
being discarded at this architectural step.

\begin{table}[h]
\centering\small
\caption{Glyph separability before and after the $2{\times}2$ patch merger.}
\label{tab:merger}
\begin{tabular}{@{}llccc@{}}
\toprule
script & checkpoint & before & after & $\Delta$ \\
\midrule
Chinese & \base{} & \num{0.707} & \num{0.700} & $-\num{0.007}$ \\
Chinese & \ours{} & \num{0.989} & \num{0.985} & $-\num{0.004}$ \\
Latin   & \base{} & \num{0.555} & \num{0.573} & $+\num{0.018}$ \\
Latin   & \ours{} & \num{0.699} & \num{0.713} & $+\num{0.014}$ \\
\bottomrule
\end{tabular}
\end{table}

\subsection{Presence-probe control}
\label{app:fingerprint}
The text-presence probe is sensitive to generation shortcuts if text-present
and text-free samples follow different synthesis paths. Bypassing the
ControlNet branch for text-free samples inflates AUC by roughly
\num{0.03}--\num{0.05}. All results in the paper therefore use matched
generation paths, with an all-black conditioning image replacing the glyph
pattern in the text-free member. Shuffled-label controls remain at
\num{0.47}--\num{0.52} AUC.

\begin{table}[h]
\centering\small
\caption{Presence-probe AUC on \ctrlpair{}. The controlled columns use the same
generation path for both members of each pair.}
\label{tab:presence}
\begin{tabular}{lcc|cc}
\toprule
& \multicolumn{2}{c|}{same pipeline} & \multicolumn{2}{c}{different pipelines} \\
block & \base{} & \ours{} & \base{} & \ours{} \\
\midrule
L6  & 0.964 & 0.977 & 0.996 & 0.996 \\
L12 & 0.977 & 0.995 & 0.995 & 0.998 \\
L18 & 0.981 & 0.995 & 0.993 & 0.996 \\
L24 & 0.967 & 0.991 & 0.991 & 0.995 \\
L27 & 0.956 & 0.991 & 0.984 & 0.996 \\
\bottomrule
\end{tabular}
\end{table}

\subsection{Glyph-identity probe after training}
\label{app:glyph}
For the before/after comparison in \S\ref{sec:why-works}, we evaluate both
checkpoints at L18, the strongest base layer shared by the two scripts.
Training raises Chinese macro-AUC from \num{0.718} to \num{0.975} and Latin
macro-AUC from \num{0.612} to \num{0.779}, while presence detection was already
near ceiling.

\begin{table}[h]
\centering\small
\caption{Glyph-identity macro-AUC at L18 before and after training.}
\label{tab:glyph-summary}
\begin{tabular}{@{}lccc@{}}
\toprule
script & \base{} & \ours{} & $\Delta$ \\
\midrule
Chinese & \num{0.718} & \num{0.975} & $+\num{0.257}$ \\
Latin   & \num{0.612} & \num{0.779} & $+\num{0.167}$ \\
\bottomrule
\end{tabular}
\end{table}

% =====================================================================
\section{Frequency and Preprocessing Controls}
\label{app:freq}

\subsection{External-benchmark and preprocessing notes}
\label{app:script}
\label{app:hcbench-human}
HC-Bench contains \num{56} hidden-text images. Two readers, evaluated
independently with squinting permitted, recovered the hidden text in
\num{35.7}\% of them on average. We therefore treat HC-Bench as an external
transfer benchmark rather than a human-legibility ceiling.

For the reduced-view rows in Table~\ref{tab:mllm}, the model preprocessor
enforces a minimum input area of \num{65536} pixels. Both the $1/8$ and
$40{\times}40$ conditions therefore reach the encoder at approximately
$256^2$ pixels and \num{64} visual tokens. The labels in Table~\ref{tab:mllm}
describe the image before this preprocessing step.

The \num{64}-token condition of \S\ref{sec:freq} is \emph{not} the same
manipulation as these rows, even though both end at \num{64} visual tokens.
Table~\ref{tab:mllm} replaces the whole image with a reduced view and then
lets the preprocessor resize it, whereas \S\ref{sec:freq} pools the original
view down to \num{64} tokens directly; the two therefore differ in spectrum
and in score (\num{29.2} versus \num{38.22} for the base model). Neither
number should be read as a replication of the other.

\subsection{\textcolor{blue}{Script-specific transcription}}
\label{app:script-transcription}
\begingroup
\color{blue}
We evaluate \base{}, SFT, and \ours{} on the full \num{1076}-image \bench{}.
All predictions use the original images and the original transcription prompt
(\texttt{format\_prompt}). For each image, we separately project the reference
and predicted transcription onto Han characters and Latin letters, preserving
their order and removing digits, punctuation, whitespace, and other characters.
We then compute case-insensitive ANLS on each projected string, using a
normalized edit-distance threshold of $0.5$; an empty prediction scores zero.
Table~\ref{tab:script-transcription} reports the mean scores across images.

These are character-projection diagnostics on \bench{}, not
evaluations on separate Chinese- and English-language datasets. In particular,
Latin-letter strings may include domains or identifiers rather than natural
English text, and projection excludes errors in the removed characters.

\begin{table}[h]
\centering\small
\caption{\textcolor{blue}{Script-specific transcription ANLS (\%) on the
full \num{1076}-image \bench{}.}}
\label{tab:script-transcription}
\begin{tabular}{@{}lcc@{}}
\toprule
Model & Han characters & Latin letters \\
\midrule
\base{} & \num{24.34} & \num{33.46} \\
SFT & \num{70.94} & \num{85.16} \\
\ours{} & \textbf{\num{84.09}} & \textbf{\num{94.20}} \\
\bottomrule
\end{tabular}
\end{table}

\paragraph{Cross-model comparison.}
To examine whether script composition may affect the model ranking, we compare
Qwen3.5-9B BASE with the commercial models on the same \num{1035} images whose
ground-truth transcriptions contain both Han characters and Latin letters.
This subset analysis is distinct from the full \num{1076}-image comparison
above. Table~\ref{tab:cross-model-script} reports Han- and Latin-projected ANLS
from the original-image predictions.
Qwen3.5-9B scores higher than Claude on Han characters
(\num{24.27} vs.\ \num{14.28}), while their Latin-letter scores are nearly
identical (\num{33.49} vs.\ \num{33.48}). This pattern suggests that script
composition may contribute to their overall ranking, rather than establishing
a general advantage over larger models. Qwen3.5-9B also scores higher than
Gemini and GPT-astra on both projections.

\begin{table}[h]
\centering\small
\caption{\textcolor{blue}{Script-projected transcription ANLS (\%) for
off-the-shelf models on the same \num{1035} mixed-script \bench{} images,
using original-image inputs.}}
\label{tab:cross-model-script}
\begin{tabular}{@{}lcc@{}}
\toprule
Model & Han characters & Latin letters \\
\midrule
Qwen3.5-9B BASE & \num{24.27} & \num{33.49} \\
Claude Opus 5 & \num{14.28} & \num{33.48} \\
Gemini 3.1 Pro & \num{15.14} & \num{29.00} \\
GPT-6-astra & \num{7.94} & \num{20.90} \\
\bottomrule
\end{tabular}
\end{table}
\endgroup

\subsection{Spectral audit of the reduced views}
\label{app:preproc}
\label{app:spectra}
Because the two reduced conditions are matched in token count, their difference
is primarily spectral. Table~\ref{tab:bands} shows that the $40{\times}40$
condition retains a larger fraction of energy in the lowest band and less
energy in the higher bands than the $1/8$ condition. This explains why nominal
scale alone is an incomplete description of the corrective view.

\begin{table}[h]
\centering\small
\caption{Share of non-DC spectral energy by band, in cycles per pixel, for the
three \bench{} viewing conditions.}
\label{tab:bands}
\begin{tabular}{@{}lccc@{}}
\toprule
band (c/p) & orig. & $1/8$ & $40{\times}40$ \\
\midrule
$0$--$1/32$    & \num{91.3}\% & \num{94.3}\% & \textbf{\num{96.5}\%} \\
$1/32$--$1/16$ & \num{6.5}\%  & \num{4.8}\%  & \num{3.3}\% \\
$1/16$--$1/8$  & \num{1.7}\%  & \num{0.87}\% & \num{0.21}\% \\
$>1/8$         & \num{0.5}\%  & \num{0.08}\% & \num{0.03}\% \\
\bottomrule
\end{tabular}
\end{table}

\subsection{Frequency at constant token count}
\begin{figure}[h]
\centering
\includegraphics[width=0.48\textwidth]{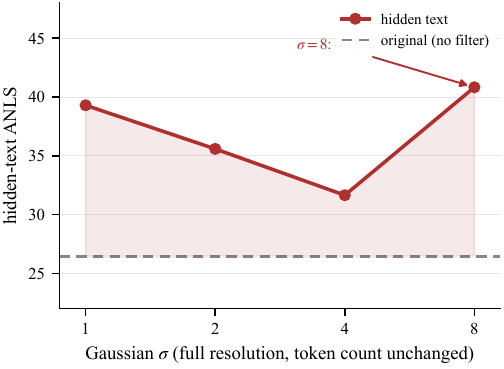}
\caption{\textbf{Frequency at constant token count.} Hidden-text reading
ANLS on all \num{1076} \bench{} images under Gaussian low-pass filtering.
The dashed line is the unfiltered input; the visual-token count is unchanged
throughout the sweep.}
\label{fig:gauss}
\end{figure}

A Gaussian low-pass applied at full resolution leaves the visual-token count
unchanged. Every measured $\sigma$ improves over the unfiltered
\num{26.44} ANLS baseline, by \num{5.2}--\num{14.4} points
(Figure~\ref{fig:gauss}). The curve is not monotonic in $\sigma$, so we use it
only to establish that suppressing high-frequency content helps independently
of sequence length.

\subsection{Contrast controls}
Contrast-only manipulation does not reproduce the frequency effect.
CLAHE \citep{PIZER1987355} leaves the spectrum approximately intact and
yields essentially no gain, whereas operations that also flatten texture, such
as bilateral filtering \citep{710815}, improve more
(Table~\ref{tab:h3}). We therefore treat contrast as an unlikely primary
explanation for the corrective-view benefit.

\begin{table}[h]
\centering\small
\caption{Contrast and band-selective controls on \bench{}, against the
unfiltered \num{26.44} ANLS baseline.}
\label{tab:h3}
\begin{tabular}{@{}llcc@{}}
\toprule
transform & what it does & ANLS & $\Delta$ \\
\midrule
CLAHE                  & raises contrast, spectrum intact      & \num{25.91} & $-\num{0.53}$ \\
bilateral filter       & flattens texture, keeps glyph edges   & \num{32.22} & $+\num{5.79}$ \\
posterise ($3$ levels) & compresses range and texture          & \num{37.21} & $+\num{10.77}$ \\
band-stop              & removes mid, keeps low $+$ high       & \num{11.22} & $-\num{15.22}$ \\
\bottomrule
\end{tabular}
\end{table}

% =====================================================================
\section{Teacher and Distillation Controls}
\label{app:teacher}

All reported default \ours{} results use a fixed teacher copied from the
initial backbone. The teacher-input ablation in Table~\ref{tab:ablation}
separates the privileged view from the ground-truth hint. The view alone reaches
\num{54.68}, above the hint-only condition at \num{41.79}; combining both
reaches \num{88.61}. A stronger control replaces the teacher's second view with
the $1/8$ downscaled view of another sample while keeping the original image,
hint, resolution, and visual-token budget unchanged. Under the paired setting,
the matched view reaches \textcolor{blue}{\num{88.61}} and the mismatched view \num{78.60}; the
gap remains through the third epoch. This isolates the benefit of
sample-aligned corrective visual context.

\textcolor{blue}{The trajectory and supervision ablation compares hard and soft
supervision under each trajectory source. With teacher-generated offline
trajectories, hard and soft supervision reach \num{68.31} and \num{77.57},
respectively; with student-generated online trajectories, they reach
\num{72.67} and \num{88.61}. Ground-truth SFT reaches \num{72.47}.
Both the trajectory source and the supervision setup differ between the
\num{68.31} and \num{88.61} configurations.}

\subsection{\textcolor{blue}{Teacher behavior under gold and student prefixes}}
\label{app:teacher-prefix}
\begingroup\color{blue}
We examine four diagnostic settings: gold-prefix transcription, the student's
empty-answer decision, its first pure-text token error, and continuation after
an error. All teacher conditions include the original image $x$; $v$ denotes
its $1/8$-scale view and $h$ the ground-truth hint.

\paragraph{Gold-prefix transcription accuracy.}
We score the same \num{35307} transcription-field tokens under the correct
preceding output tokens, excluding JSON-format tokens. At every position,
the teacher receives the gold prefix regardless of its predictions at earlier
positions. Table~\ref{tab:teacher-gold-prefix} reports top-1 agreement with the
ground-truth token, mean ground-truth-token probability, and mean negative
log-likelihood (NLL).

\begin{table}[h]
\centering\small
\caption{\textcolor{blue}{Teacher predictions on \num{35307} gold-prefix transcription tokens; JSON-format tokens are excluded.}}
\label{tab:teacher-gold-prefix}
\begin{tabular}{@{}lccc@{}}
\toprule
Teacher input & Top-1 (\%) & GT probability (\%) & NLL \\
\midrule
$x+h$ (hint only) & \num{98.13} & \num{95.03} & \num{0.0907} \\
$x+v$ (view only) & \num{76.18} & \num{69.63} & \num{1.6323} \\
$x+v+h$ (view + hint) & \num{95.95} & \num{94.11} & \num{0.1110} \\
\bottomrule
\end{tabular}
\end{table}

\paragraph{Student-prefix empty-answer decisions.}
We sample outputs from the initial \base{} student on \num{256} positive
images. The student outputs \texttt{[]} on \num{123} images whose ground-truth
answers are non-empty. For these selected cases, we score all three teachers
at the same actual student prefix immediately before the empty-answer decision.
Table~\ref{tab:teacher-empty-top1} reports top-1 agreement with the ground-truth
opening token \texttt{[}, which begins the non-empty answer at this decision
point. Separately, the empty-answer branch is ranked first by the hint-only
teacher in \num{106}/\num{123} cases and by the view-plus-hint teacher in
\num{5}/\num{123} cases. These empty-branch counts are distinct from the
ground-truth-token counts in the table. All rates are conditional on the selected
student empty-answer subset, not population-level rates.

\begin{table}[h]
\centering\small
\caption{\textcolor{blue}{Ground-truth opening-token top-1 agreement before the empty-answer decision, on \num{123} selected student-empty cases from \num{256} positive images. All \num{123} have non-empty ground truth. Counts are shown in parentheses.}}
\label{tab:teacher-empty-top1}
\begin{tabular}{@{}lc@{}}
\toprule
Teacher input & GT top-1 (\%) \\
\midrule
$x+h$ (hint only) & \num{10.57} (13/123) \\
$x+v$ (view only) & \num{76.42} (94/123) \\
$x+v+h$ (view + hint) & \num{86.18} (106/123) \\
\bottomrule
\end{tabular}
\end{table}

\paragraph{First transcription-token error.}
We additionally inspect \num{28} cases at the first erroneous pure-text token,
excluding JSON-format and punctuation tokens. Before this first error, the
student prefix remains aligned with the gold output. Each teacher is scored
at the same prefix immediately before the erroneous token.
Table~\ref{tab:teacher-first-error-top1} reports ground-truth-token top-1
agreement.

\begin{table}[h]
\centering\small
\caption{\textcolor{blue}{Ground-truth-token top-1 agreement immediately before the first pure-text token error, on \num{28} gold-aligned student prefixes. JSON-format and punctuation tokens are excluded.}}
\label{tab:teacher-first-error-top1}
\begin{tabular}{@{}lc@{}}
\toprule
Teacher input & GT top-1 (\%) \\
\midrule
$x+h$ (hint only) & \num{67.86} (19/28) \\
$x+v$ (view only) & \num{17.86} (5/28) \\
$x+v+h$ (view + hint) & \num{71.43} (20/28) \\
\bottomrule
\end{tabular}
\end{table}

\paragraph{Continuation after an error.}
We compare continuation ANLS on the same \num{235} post-error prefixes.
Table~\ref{tab:teacher-post-error} reports both teachers' scores and their
paired difference, with a paired bootstrap 95\% confidence interval.

\begin{table}[h]
\centering\small
\caption{\textcolor{blue}{Continuation ANLS on \num{235} post-error prefixes. The difference is view-plus-hint minus hint-only; its interval is a paired bootstrap 95\% confidence interval.}}
\label{tab:teacher-post-error}
\begin{tabular}{@{}lcc@{}}
\toprule
Teacher input / comparison & ANLS / difference & 95\% CI \\
\midrule
$x+h$ (hint only) & \num{33.25} & --- \\
$x+v+h$ (view + hint) & \num{32.46} & --- \\
Difference & $-\num{0.79}$ & $[-\num{2.74},\num{1.07}]$ \\
\bottomrule
\end{tabular}
\end{table}

\paragraph{Analysis.}
The hint-only teacher accurately predicts transcription tokens under gold
prefixes, while the view-only teacher provides weaker token-level supervision.
On the selected student-empty cases, however, the hint-only teacher often
supports the impending empty-answer decision, whereas the full teacher more
often prefers the non-empty ground-truth answer. This difference in supervision
on student-generated prefixes offers an explanation for the weaker
hint-only distilled student relative to SFT, which directly supervises the gold
output. At the first pure-text error, the full and hint-only teachers differ
by only one correct prediction (\num{20}/\num{28} vs.\ \num{19}/\num{28}).
Post-error continuation likewise shows no clear advantage for the full teacher:
the ANLS difference is $-\num{0.79}$ and its confidence interval includes zero.
Across these diagnostics, the difference between the two teachers is most
evident before the student commits to an empty answer.
\endgroup

\subsection{Corrective-view and training variants}
\label{app:teacher-variants}
In all experiments, the teacher receives the same original image $x$, task
instruction $q$, and ground-truth hint $h$; we vary only the corrective view
$v$ (Table~\ref{tab:teacher-variants}). Replacing the default $1/8$ view with
a fixed $40{\times}40$ view retains nearly identical performance
(\num{88.09} vs.\ \num{88.61}), while $v=1/4$ reaches \num{78.51}. We also
tested whether complementary scales could be combined by averaging or
selecting among teacher distributions from $v\in\{1/8,\,1/4,\,40{\times}40\}$;
none improves over the single default teacher, with probability averaging
reaching \num{83.05}.

\begin{table}[h]
\centering\small
\caption{Teacher-view variants. All \ours{} teachers receive $(x,v,q,h)$;
only the corrective view $v$ or its aggregation is varied. Student inference
always uses $(x,q)$ only.}
\label{tab:teacher-variants}
\begin{tabular}{@{}llc@{}}
\toprule
teacher configuration & corrective view $v$ & \bench{} \\
\midrule
default               & $1/8$                      & \textbf{\num{88.61}} \\
fixed view            & $40{\times}40$             & \num{88.09} \\
fixed view            & $1/4$                      & \num{78.51} \\
probability mean      & $1/8,\ 1/4,\ 40{\times}40$ & \num{83.05} \\
max-entropy selection & $1/8,\ 1/4,\ 40{\times}40$ & \num{80.51} \\
min-entropy selection & $1/8,\ 1/4,\ 40{\times}40$ & \num{74.30} \\
\bottomrule
\end{tabular}
\end{table}

Beyond the view itself, updating the teacher by EMA (rate \num{0.05}) reaches
\num{79.12}. SFT warm starts reach \num{82.85} after one epoch and
\num{81.35} after two epochs, both below direct \ours{} at \num{88.61}. One
plausible explanation is that the supervised stage first concentrates
probability on a one-hot target, reducing the uncertainty over visually
plausible alternatives that \ours{} is designed to transfer. Starting with
distributional supervision preserves this structure throughout training.

We initially hypothesized that corrective views at different scales might
provide complementary teacher signals. We therefore combine the $1/8$,
$1/4$, and $40{\times}40$ teachers using probability averaging, minimum-entropy
selection, and maximum-entropy selection. None exceeds the best single fixed
teacher: probability averaging reaches \num{83.05}, while maximum- and
minimum-entropy selection reach \num{80.51} and \num{74.30}, respectively.
We think there are two reasons for this. First, the $1/8$ and $40{\times}40$
teachers are individually strong (\num{88.61}, \num{88.09}) while the $1/4$
teacher is much weaker (\num{78.51}); averaging their distributions therefore
lets the weak $1/4$ teacher drag down the two strong ones, instead of
complementing them. Second, min-entropy selection (\num{74.30}) does notably
worse than max-entropy selection (\num{80.51}): picking the most confident
teacher discards exactly the uncertainty that a distributional target is
meant to preserve --- the spread over visually confusable glyphs --- whereas
retaining more uncertainty keeps more of the alternative candidates alive
for the student to learn from.
% =====================================================================
\section{Recoverability Boundary}
\label{app:strength}

\ctrlstr{} varies only the ControlNet conditioning strength over the same
\num{200} scenes. Both \base{} and \ours{} improve as the glyph signal becomes
more visible, but the largest useful margin appears in the partially recoverable
regime. At strength \num{0.5}, where strokes are largely destroyed,
\base{} and \ours{} are statistically indistinguishable
(\num{0.597} vs.\ \num{0.615}, 95\% CI $[-0.008,+0.063]$; paired bootstrap,
\num{100} resamples). At strength
\num{0.9}, \ours{} improves from \num{0.886} to \num{0.944}; at
\num{1.2}, both models are near ceiling. This empirically delineates the recoverability boundary of the method:
training can reorganize partially recoverable evidence, but does not reconstruct
glyph structure that is largely absent from the input.

\begin{figure}[h]
\centering
\includegraphics[width=0.5\textwidth]{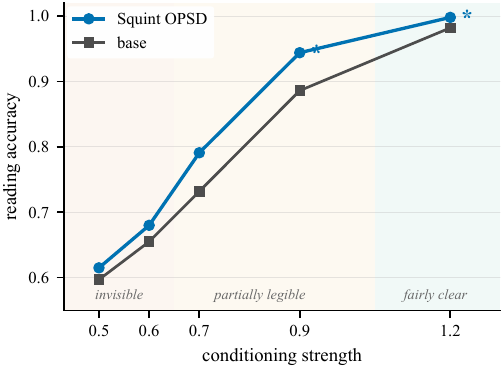}
\caption{\textbf{Recoverability versus hiding strength.} Reading accuracy for
the base model and \ours{} as conditioning strength varies, over the same
\num{200} scenes. Stars mark strength levels at which the 95\% CI of the
paired base-vs-\ours{} difference excludes zero (paired bootstrap,
\num{100} resamples); unmarked levels are not significant at this threshold,
and \num{0.7} is borderline and is left unmarked. The method provides
its clearest benefit in the partially recoverable regime; the two ends
saturate for different reasons.}
\label{fig:strength}
\end{figure}

% =====================================================================
\section{Analysis Details}

\subsection{Candidate-distribution metrics}
\label{app:runnerup-metrics}
For \S\ref{sec:runnerup}, we analyze \num{200} \ctrlread{} images. We retain
examples whose ground-truth characters are each represented by a single
token, yielding \num{367} aligned glyph positions before candidate filtering.
At each position, we run the model under teacher forcing and take its top-$20$
next-token candidates. We remove the ground-truth token, retain only tokens
that decode to exactly one CJK codepoint, and keep positions with at least five
valid candidates. Thus, the analysis concerns non-target high-probability
alternatives rather than the full vocabulary tail.

For $\rho_{\mathrm{vis}}$, each candidate and target glyph is rendered with
Noto Sans CJK SC, centered by its ink bounding box, dilated with a
$3{\times}3$ max filter, resized to $64{\times}64$ with Lanczos resampling
\citep{doi:10.1137/0701007}, and binarized. Visual similarity is pixel IoU. We compute a per-position
Spearman correlation between candidate probability and glyph IoU, subtract the
mean correlation from \num{20} random permutations of the similarity scores,
and average over valid positions.

For $\rho_{\mathrm{lang}}$, we evaluate the same checkpoint without the image,
using the identical OCR prompt and teacher-forced ground-truth prefix. We then
compute the per-position Spearman correlation between the image-conditioned
candidate probabilities and their probabilities under this matched text-only
context. Multi-token ground-truth characters are excluded from the analysis.

\subsection{Matched-position robustness}
\label{app:runnerup-common}
The default filtering retains different numbers of valid positions for the
three checkpoints because their top-$20$ distributions have different
sharpness: \num{254} for \base{}, \num{259} for SFT, and \num{164} for
\ours{}. To remove this selection effect, the main-text comparison uses the
intersection of valid positions across all three models ($n=\num{120}$).
Table~\ref{tab:runnerup-common} shows that the ordering is unchanged on this
matched set, so the main-text comparison is not an artefact of position
selection. The same conclusion
holds when intersecting all five analyzed conditions ($n=\num{106}$), where
$\rho_{\mathrm{vis}}$ is \num{0.117}/\num{0.247}/\num{0.298} for
\base{}/SFT/\ours{}, respectively.

\begin{table}[h]
\centering\small
\caption{Candidate-distribution structure before and after matching glyph
positions across checkpoints. Parentheses give the number of valid positions.}
\label{tab:runnerup-common}
\begin{tabular}{@{}lccc@{}}
\toprule
metric & \base{} & SFT & \ours{} \\
\midrule
$\rho_{\mathrm{vis}}$, own valid set
  & \num{0.079} (254) & \num{0.160} (259) & \num{0.243} (164) \\
$\rho_{\mathrm{vis}}$, common set
  & \num{0.126} & \num{0.242} & \textbf{\num{0.286}} \\
$\rho_{\mathrm{lang}}$, common set
  & \num{0.555} & \num{0.334} & \num{0.544} \\
\bottomrule
\end{tabular}
\end{table}

\subsection{Candidate-set sensitivity}
\label{app:runnerup-k}
Because a fixed top-$K$ cutoff interacts with distribution sharpness, we sweep
$K\in\{10,12,15,20,25,30,40,50\}$ and recompute the matched-position
comparison at each cutoff. The ordering $\ours{}>\mathrm{SFT}>\base{}$ in
$\rho_{\mathrm{vis}}$ is stable for $K=10$--$40$. At $K=50$, SFT exceeds
\ours{} (\num{0.170} vs.\ \num{0.144}; \base{}=\num{0.098}). This
sensitivity is consistent with the substantially sharper \ours{} distribution:
its top-$20$ contains fewer valid single-CJK alternatives on average
(\num{4.6}, versus \num{8.4} for \base{} and \num{7.8} for SFT), while its
target confidence is substantially higher. Expanding to a large fixed $K$
therefore adds a larger low-probability tail relative to its effective support.
We accordingly scope the main claim to the organization of high-probability
candidate alternatives rather than the entire vocabulary tail.

\subsection{Attention-mass definition}
\label{app:amr}
For each answer glyph position $t\in\mathcal{A}$, we renormalize attention over
the visual tokens $\mathcal{V}$ and measure the mass falling on the
glyph-bearing set $\mathcal{T}$, divided by its chance level:
\begin{equation}
\amr_{\text{norm}}
=
\frac{1}{|\mathcal{A}|}
\sum_{t\in\mathcal{A}}
\sum_{j\in\mathcal{T}}\tilde a_{t,j}
\;\Big/\;
\frac{|\mathcal{T}|}{|\mathcal{V}|},
\qquad
\tilde a_{t,j}
=
\frac{a_{t,j}}{\sum_{j'\in\mathcal{V}}a_{t,j'}} .
\end{equation}
Here $\mathcal{A}$ includes only answer positions corresponding to transcribed
glyphs, excluding formatting tokens. The known glyph mask is downsampled
through the same $2{\times}2$ merger as the visual features before attention is
measured. Thus $\amr_{\text{norm}}=1$ denotes chance-level allocation within
the visual tokens.

\subsection{Visual-token embedding knockout}
\label{app:knockout}
To test whether reading causally depends on glyph-bearing visual regions, we
intervene on the output embeddings of the vision tower before they are passed
to the language model. The intervention is applied once during prefill; no
language-model attention logits or weights are modified.

Let $T$ denote visual tokens whose receptive fields overlap the known glyph
pattern. For the glyph knockout, each embedding in $T$ is replaced by the mean
embedding of the non-glyph tokens from the same image. This preserves sequence
length, token positions, and the subsequent attention computation. As a
matched control, we sample the same number of tokens uniformly from the
non-glyph region using a fixed random seed and apply the identical
mean-embedding replacement.

We additionally dilate the glyph mask by $d\in\{0,1,2\}$ iterations of a
$3{\times}3$ neighborhood operation. These interventions replace on average
\num{169.0} (\num{16.5}\%), \num{299.9} (\num{29.3}\%), and \num{391.8}
(\num{38.3}\%) of the \num{1024} visual tokens, respectively. Generation is
greedy with a maximum of \num{64} new tokens, and a prediction is counted as
correct when the ground-truth string occurs in the generated transcription
after whitespace removal.

Table~\ref{tab:knockout} shows that the effect is strongly spatially specific.
At $d=2$, \ours{} decreases from \num{90.5}\% to \num{71.5}\%, while the
matched random intervention changes accuracy only from \num{90.5}\% to
\num{91.0}\%. \base{} also degrades under glyph removal (\num{26.0}\% to
\num{15.0}\%), but begins from substantially lower reading accuracy.

\begin{table}[h]
\centering\small
\caption{Reading accuracy after visual-token embedding replacement.
Glyph-region embeddings are replaced by the mean embedding of non-glyph
tokens from the same image; the random control replaces an equal number of
randomly selected background-token embeddings using the same operation. $d$
denotes the number of $3{\times}3$ dilations applied to the known glyph
mask.}
\label{tab:knockout}
\begin{tabular}{lcc}
\toprule
intervention & \base{} & \ours{} \\
\midrule
none (baseline)                & 26.0 & 90.5 \\
replace glyph tokens, $d{=}0$  & 19.5 & 87.0 \\
replace glyph tokens, $d{=}1$  & 16.0 & 77.0 \\
replace glyph tokens, $d{=}2$  & 15.0 & 71.5 \\
\midrule
replace random tokens, $d{=}0$ & 25.0 & 91.5 \\
replace random tokens, $d{=}1$ & 25.0 & 91.0 \\
replace random tokens, $d{=}2$ & 24.5 & 91.0 \\
\bottomrule
\end{tabular}
\end{table}

% =====================================================================
\section{Training and Reproducibility}
\label{app:trainmix}

\subsection{Training mixture}
\begin{table}[h]
\centering\small
\caption{Training mixture, \num{5905} prompts after removing \num{65} low-quality samples. DocVQA contributes train-split samples only; its
reported evaluation uses a disjoint test split.}
\label{tab:trainmix}
\begin{tabular}{llr}
\toprule
group & sources & prompts \\
\midrule
\textcolor{blue}{synthetic text / no-text}
 & AIGC collection 1143, self-synthesised ControlNet 1144, & \\
 & FakeTextDetect 1267, minus 65 removed & 3489 \\
open-source OCR
 & TextCaps 564, DocVQA 302, CASIA 150, RCTW 136, & \\
 & POIE 56, FUNSD 3, ICDAR13 2 & 1213 \\
VQA
 & OCR-VQA & 1203 \\
\midrule
\multicolumn{2}{l}{total} & 5905 \\
\bottomrule
\end{tabular}
\end{table}

\noindent Dataset references: TextCaps \citep{10.1007/978-3-030-58536-5_44}, DocVQA
\citep{Mathew_2021_WACV}, CASIA \citep{6065272}, RCTW
\citep{8270164}, FUNSD \citep{8892998}, ICDAR13
\citep{karatzas2013icdar}, and OCR-VQA \citep{mishraICDAR19}.

\textcolor{blue}{The AIGC collection uses QR Monster with Stable Diffusion~1.5, while the self-synthesised branch uses QR Monster~2; both use ControlNet conditioning. The \num{1267} FakeTextDetect examples have empty-list targets (\texttt{[]}) and serve as no-text negatives, not hidden-text transcription examples. The per-source counts in Table~\ref{tab:trainmix} precede removal of the \num{65} low-quality samples.}

\textcolor{blue}{Synthetic training images use randomly composed text and single-line or multi-line layouts, with background prompts sampled from JourneyDB. These training images are constructed separately from the production-derived \bench{} evaluation images. The complete text strings used for synthetic training do not overlap with those in \bench{}. The synthetic training images and \probeset{} use different background pools and separately constructed conditioning images. In contrast to the broader training text, \ctrlglyph{} uses random combinations of two or three characters from fixed Chinese and Latin vocabularies.}

% TODO(author): report the image-level exact/near-duplicate audit before claiming verified deduplication.
\subsection{Held-out validation mixture}
\label{app:valmix}
We reserve a held-out split of \num{1045} examples drawn from the same source
mixture as the training data: \num{616} hidden-text examples, \num{269}
open-source OCR examples, and \num{160} region-conditioned OCR/VQA examples.
The split is separate from the \num{5905} training prompts and is intended for
model-selection and validation under the same task mixture as training.

% TODO(author): the currently reported checkpoint must be re-selected using this
% held-out split before stating that checkpoint selection was performed on it.
% Also document the actual image/prompt/source-level deduplication rules used
% between train/validation and any near-duplicate filtering, if applicable.

\subsection{Optimization and rollout settings}
\label{app:training}
The default \ours{} run initializes the student from Qwen3.5-9B
\citep{qwen35} and uses a fixed teacher copied from the same initial backbone. All student parameters,
including the vision tower, are updated. Training is configured for up to three
epochs, with checkpoints evaluated every \num{10} optimization steps.

\todo{Before submission, re-select the reported checkpoint on the held-out
validation split in Appendix~\ref{app:valmix}, then state the final selection
rule here.}

\begin{table}[h]
\centering\small
\caption{Main \ours{} optimization and rollout settings.}
\label{tab:hyper}
\begin{tabular}{@{}ll@{}}
\toprule
hyperparameter & value \\
\midrule
optimizer / learning rate & AdamW \citep{loshchilov2018decoupled}, $1{\times}10^{-5}$ \\
schedule / warmup & constant / 10 steps \\
weight decay / grad clip & \num{0.01} / \num{1.0} \\
prompt batch & \num{56} prompts per step \\
rollouts per prompt & $n{=}\num{8}$ (\num{448} sequences/step) \\
inner epochs & \num{1} \\
maximum training horizon & \num{3} epochs \\
teacher & fixed initial-backbone copy \\
trainable parameters & full model, including vision tower \\
distillation support & top-$K{=}\num{100}$ with tail bucket \\
JSD & symmetric, $\alpha=\num{0.5}$ \\
max prompt / response length & \num{8000} / \num{1024} tokens \\
rollout sampling & $T{=}\num{1.0}$, top-$p{=}\num{1.0}$ \\
validation decoding & greedy ($T{=}\num{0}$, $n{=}\num{1}$) \\
precision / training stack & bf16, FSDP \citep{10.14778/3611540.3611569}, gradient checkpointing \\
checkpoint evaluation & every \num{10} steps \\
\bottomrule
\end{tabular}
\end{table}

Student rollouts are sampled from the current policy and treated as
stop-gradient trajectories. At each sampled prefix, the teacher and student
distributions are evaluated on the student's top-100 support plus one tail
bucket containing the remaining probability mass. Generation terminates at EOS
or the maximum response length, and gradients are propagated only through the
student logits.

\paragraph{Supervised and RL baselines.}
SFT updates the full model for \num{3} epochs with learning rate
$1{\times}10^{-5}$, cosine decay, warmup ratio \num{0.03}, effective batch size
\num{64}, and maximum sequence length \num{2048}. DAPO \citep{yu2025dapoopensourcellmreinforcement},
which extends the clipped policy-gradient objective of GRPO
\citep{shao2024deepseekmathpushinglimitsmathematical}, is also trained for up to \num{3} epochs. It
uses learning rate $1{\times}10^{-5}$, rollout $n{=}\num{8}$, clipping
$(\epsilon_{\mathrm{low}},\epsilon_{\mathrm{high}})
=(\num{0.2},\num{0.28})$, and ANLS reward with threshold $\tau=\num{0.5}$.

\subsection{Visual adaptation under supervised fine-tuning}
\label{app:sft-modules}
To test whether the importance of visual adaptation is specific to the OPSD
objective, we repeat supervised fine-tuning while changing which modules are
trainable. With the vision tower frozen, SFT reaches only \num{27.29} on
\bench{}; updating both the vision tower and language model raises this to
\num{72.47}. The same qualitative dependence therefore appears under a
standard supervised objective.

\begin{table}[h]
\centering\small
\caption{Effect of visual adaptation under three-epoch supervised fine-tuning.}
\label{tab:sft-modules}
\begin{tabular}{@{}lcccc@{}}
\toprule
trainable modules & \bench{} & AI2D & DocVQA & OCRBench \\
\midrule
LM only; ViT frozen & \num{27.29} & \num{86.66} & \num{50.39} & \num{741} \\
ViT + LM            & \num{72.47} & \num{85.52} & \num{64.73} & \num{722} \\
\bottomrule
\end{tabular}
\end{table}

\subsection{Training prompts}
\label{app:prompts}
The student receives the original image and the task instruction. The teacher
receives the original image together with its $1/8$ downscaled view, followed
by the same task instruction and output format; the ground-truth output is
provided only as privileged teacher context.

\paragraph{Student template.}
\begin{quote}\small
\texttt{<image>}\\
\texttt{\{task instruction\}}\\
\texttt{\{task-specific format instruction\}}
\end{quote}

\paragraph{Teacher template.}
\begin{quote}\small
\texttt{<image>}\\
\texttt{<image>}\\
\zh{第一张图为原图,第二张图为同一张图的 1/8 下采样视图。}\\
\texttt{\{task instruction\}}\\
\texttt{\{task-specific format instruction\}}\\
\zh{该图的正确输出为\{gt\}。}
\end{quote}

For OCR-style samples, the output format is a JSON list of
\texttt{\{"text\_content": "..."\}} objects, with \texttt{[]} used when no text
is present. OCR-VQA and detection samples use the same schema with
task-specific wording.

\subsection{\probeset{} synthesis}
\label{app:synthesis}
The main synthetic branch uses Stable Diffusion 1.5 with QR Code Monster-style
conditioning. Images are generated at $512$ pixels and upscaled to $1024$, with
ControlNet scale \num{1.2}, two-stage CFG \num{5.0}/\num{7.5}, and img2img
strength \num{0.5}. We use seed \num{42} and generate \num{10} samples per
prompt. The conditioning image is retained to derive the exact glyph mask used
by the controlled diagnostic experiments.

% The examples are single-page PDF figures directly under figures/ (2026-10-09):
% on the arXiv upload path every RASTER image (png/jpg) was silently dropped
% (3 attempts, different names/paths/sizes) while PDF figures always arrived,
% so the 12 example images are now wrapped as PDFs. PNG/JPG originals are kept
% in figures/A.appendix/{ctrl_examples,show}/ -- nothing else changed.
\newcommand{\ctrlimg}[1]{\includegraphics[width=0.30\textwidth]{figures/#1}}
% Row labels lifted to the vertical middle of the (square) images.
\newcommand{\ctrlrowlabel}[1]{\raisebox{\dimexpr0.15\textwidth-0.7ex\relax}{#1}}
\begin{figure}[t]
\centering
\setlength{\tabcolsep}{6pt}
\begin{tabular}{@{}rcc@{}}
\ctrlrowlabel{\ctrlpair{}}  & \ctrlimg{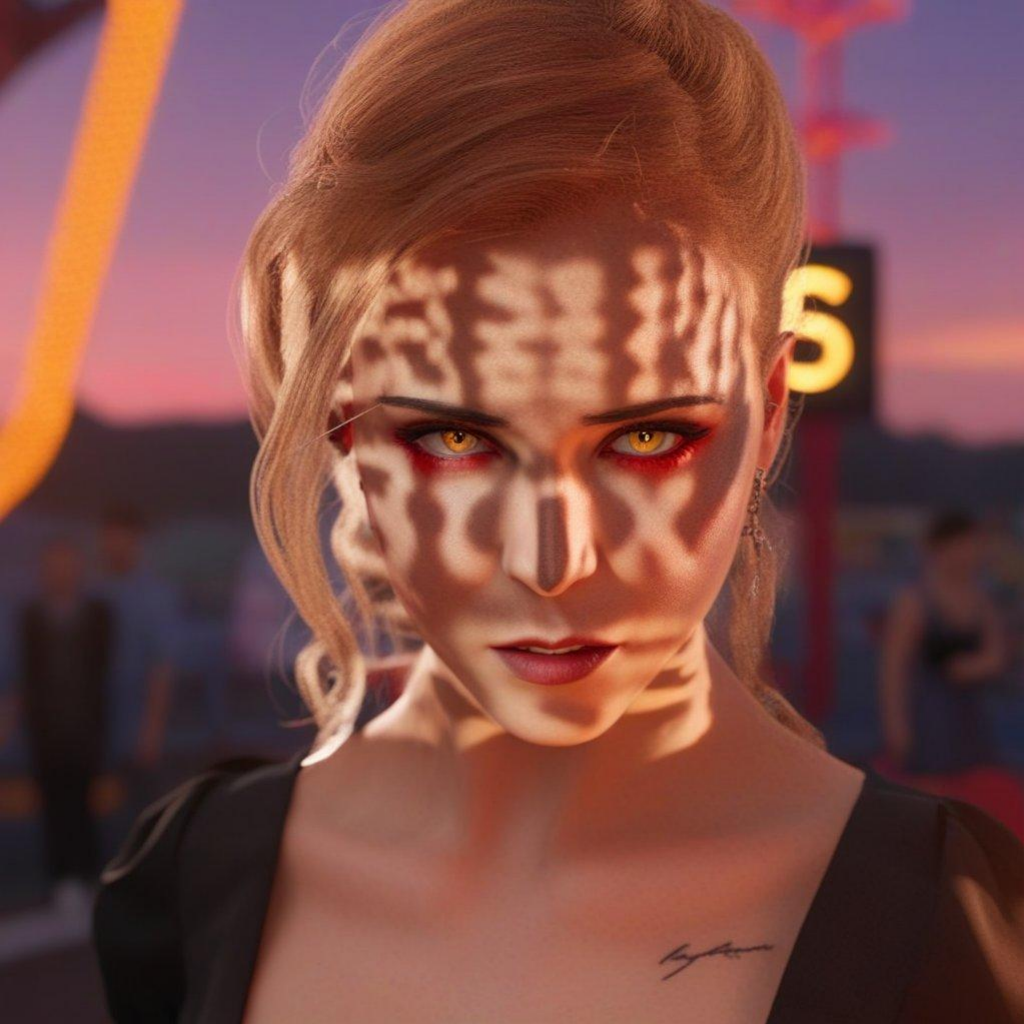}     & \ctrlimg{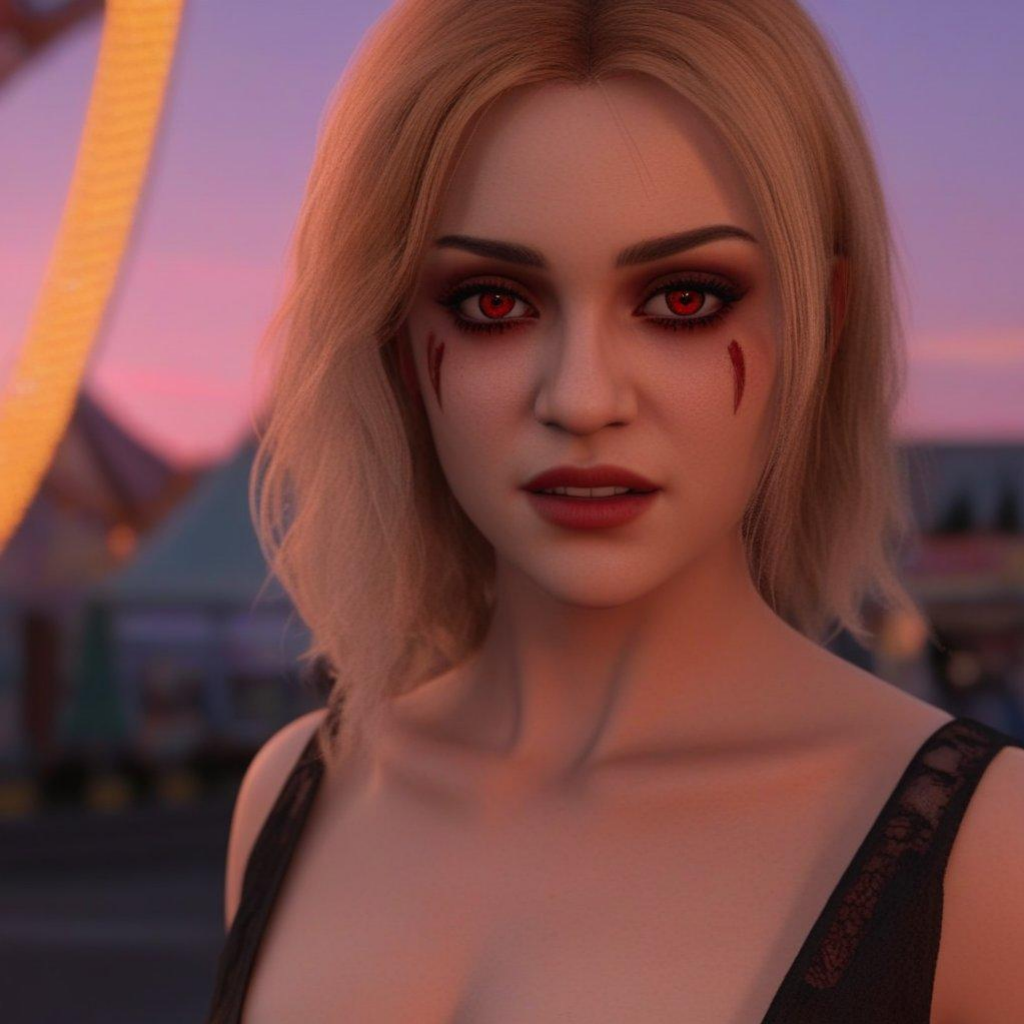} \\
              & text-present              & text-free (regenerated) \\[2pt]
\ctrlrowlabel{\ctrlglyph{}} & \ctrlimg{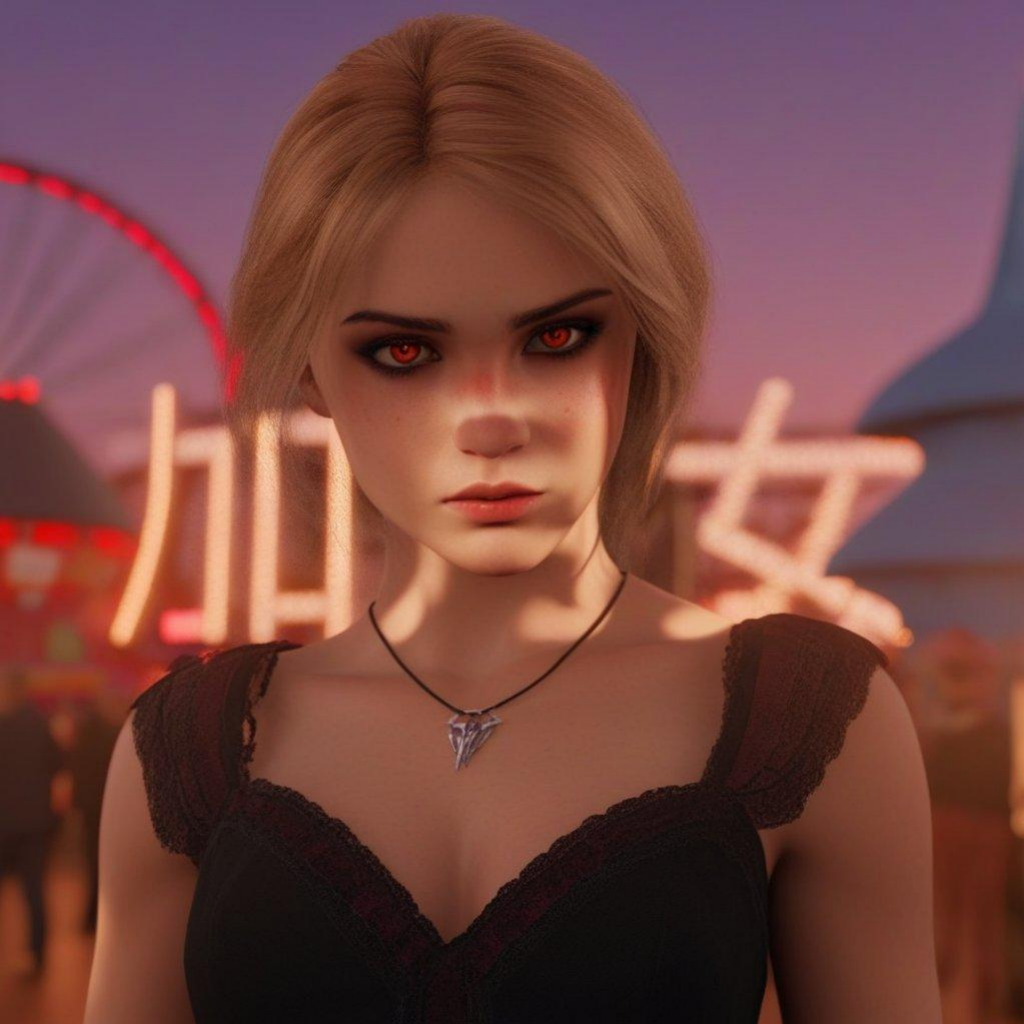}     & \ctrlimg{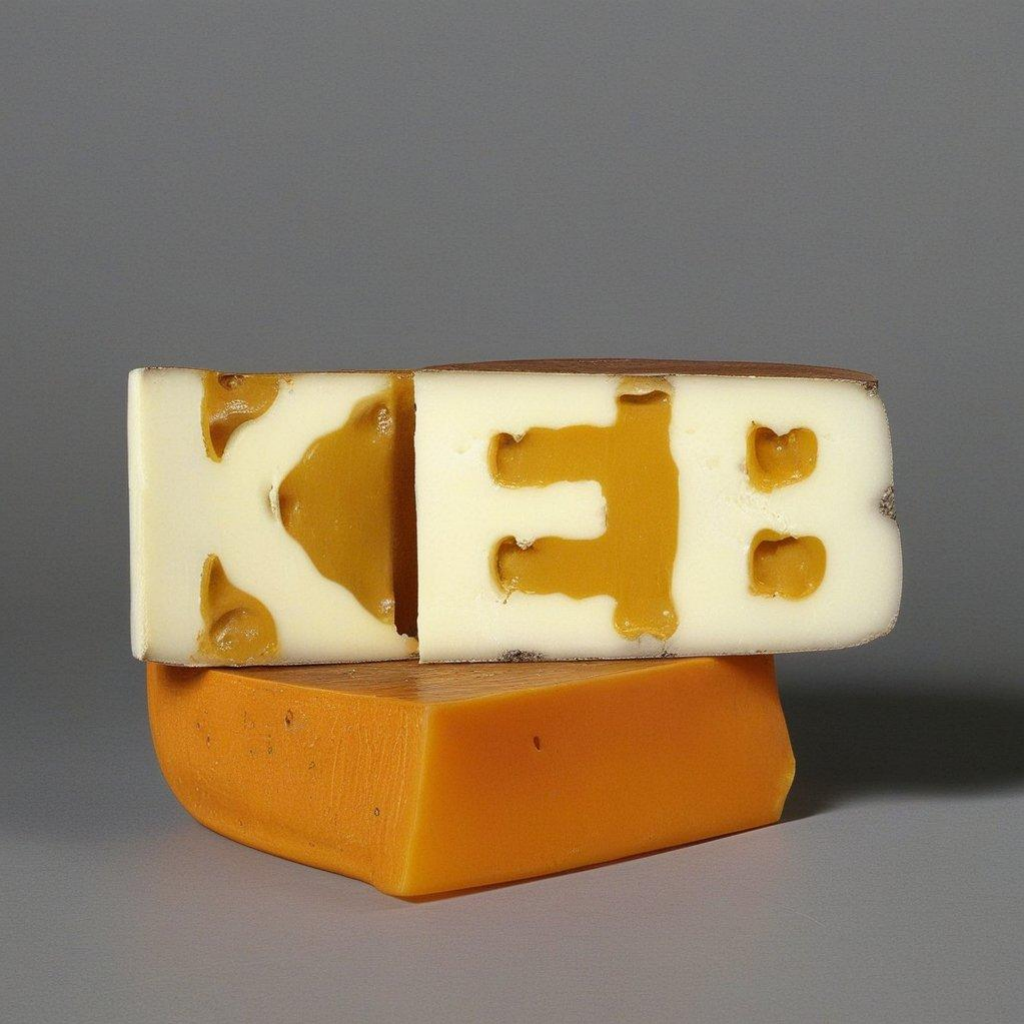} \\
              & Chinese vocabulary        & Latin vocabulary \\[2pt]
\ctrlrowlabel{\ctrlstr{}}   & \ctrlimg{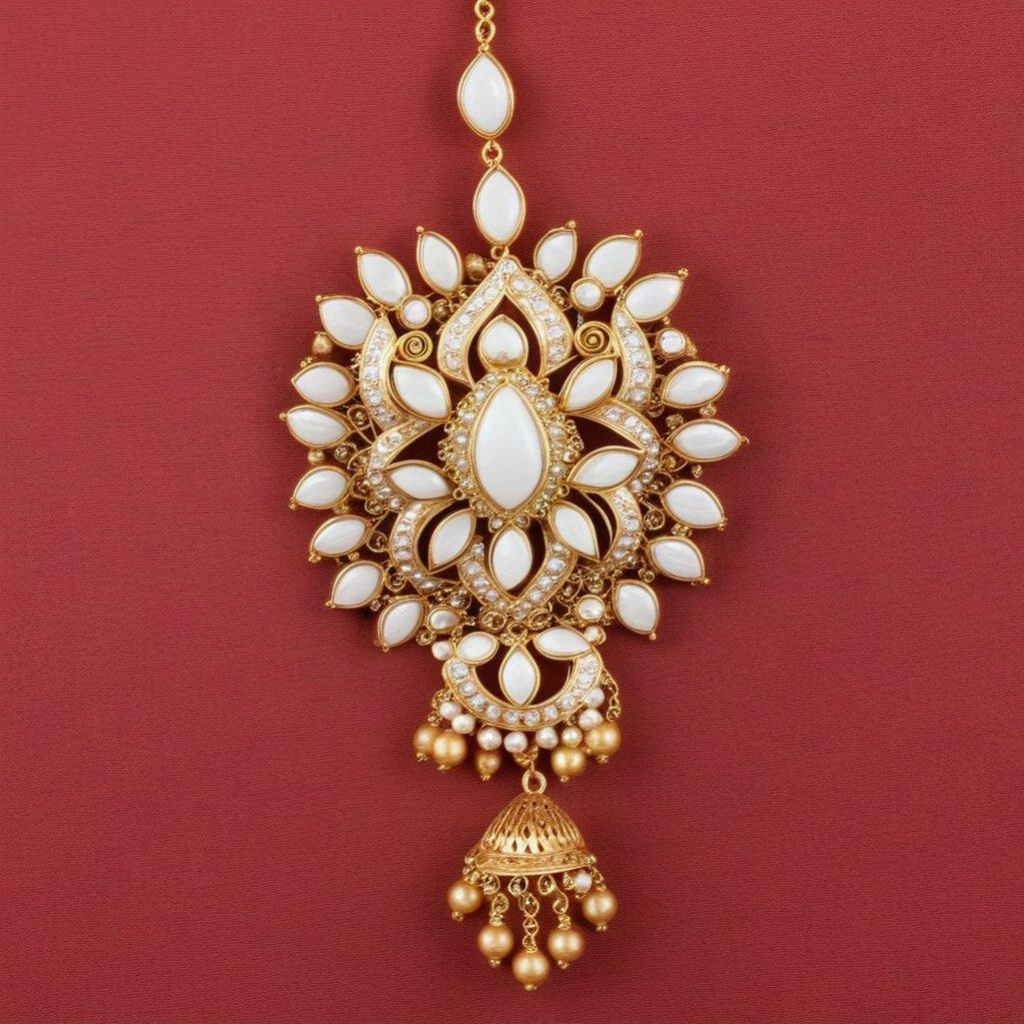}  & \ctrlimg{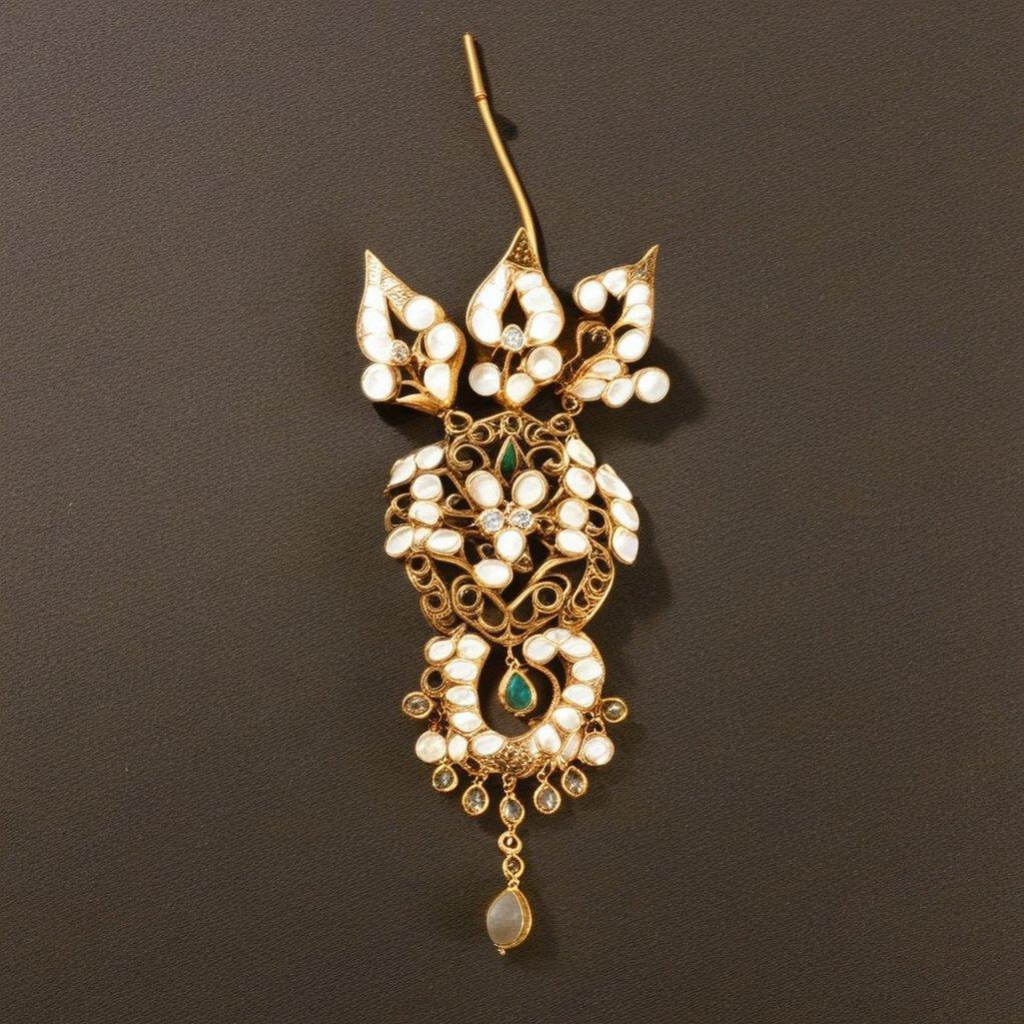} \\
              & weak conditioning         & strong conditioning \\[2pt]
\ctrlrowlabel{\ctrlread{}}  & \ctrlimg{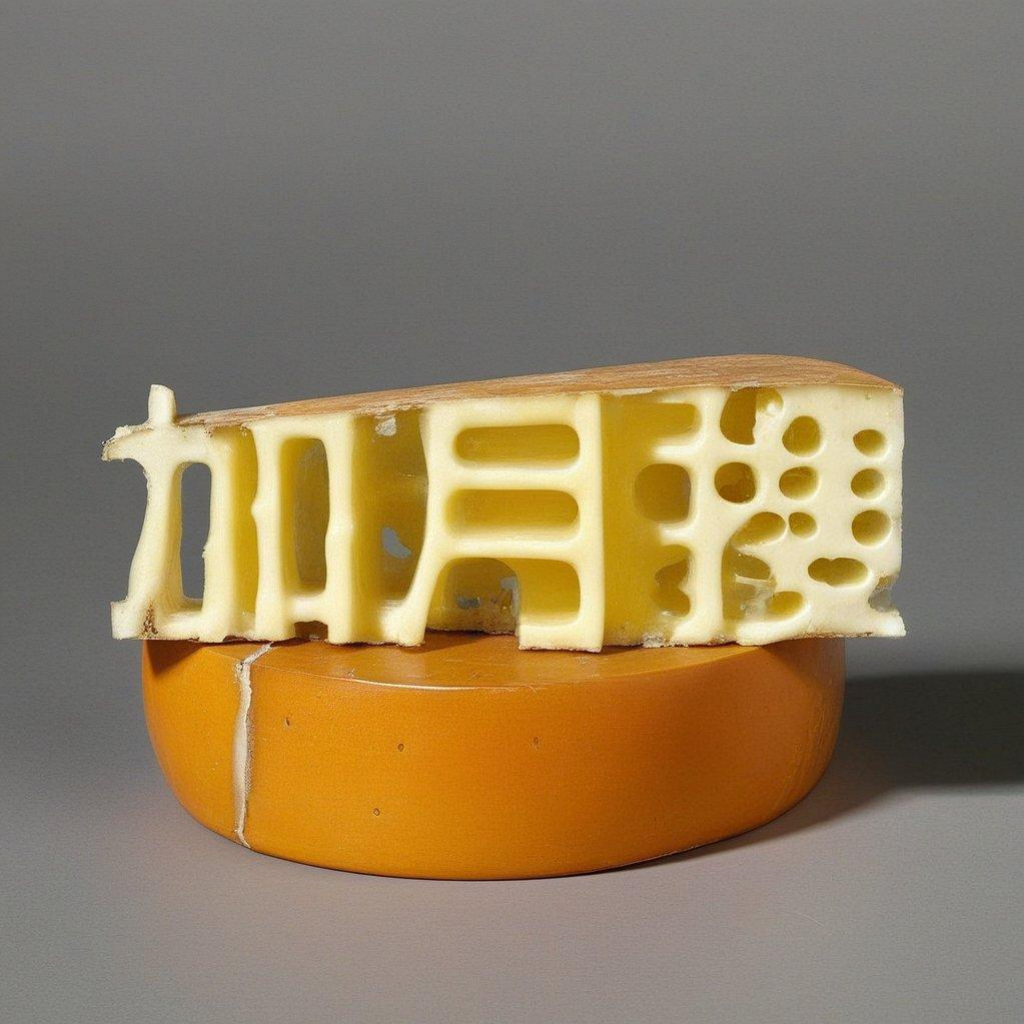}        & \ctrlimg{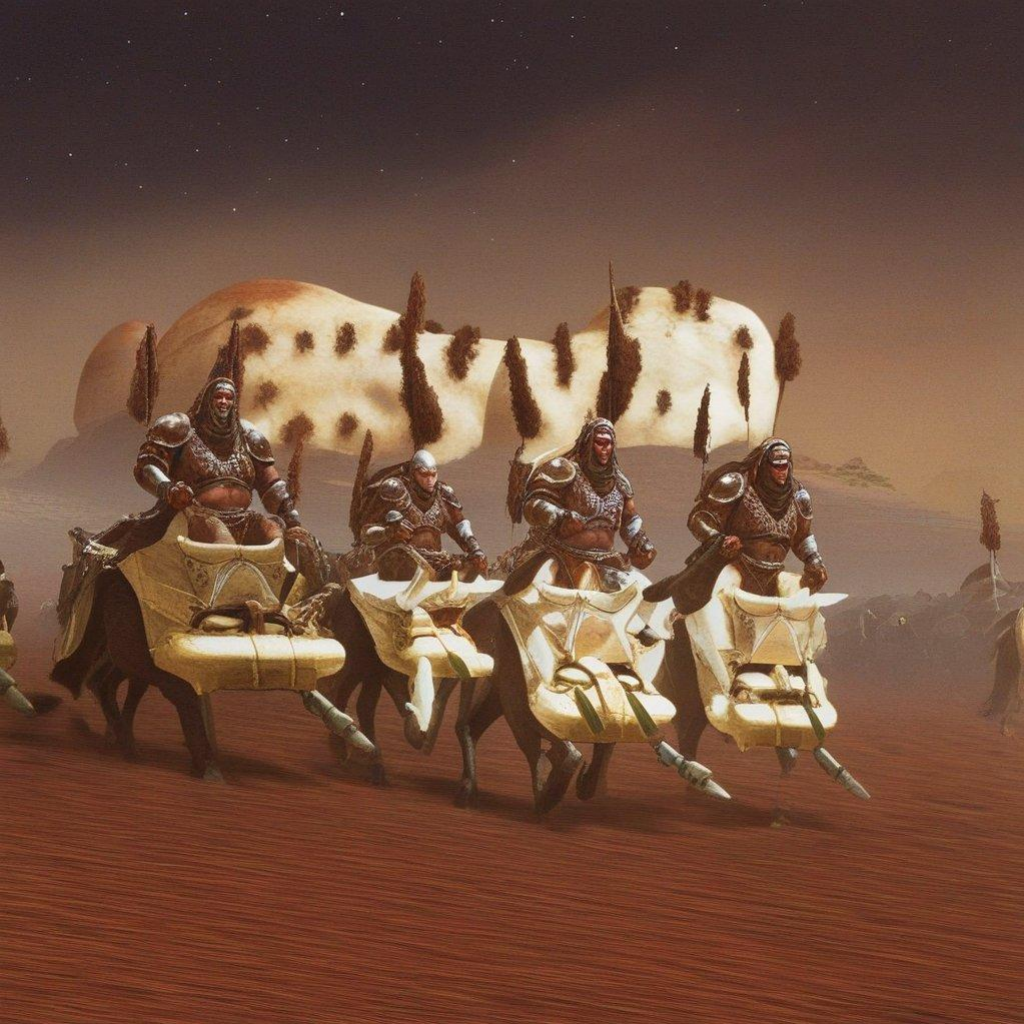} \\
              & \multicolumn{2}{c}{open-vocabulary read-out} \\
\end{tabular}
\caption{Two examples from each of the four \probeset{} subsets. The embedded
strings are carried in the images' coarse structure: they are hard to notice
at full size but become legible when the page is squinted at or viewed at
reduced scale. \ctrlpair{}: a text-present image and its text-free
counterpart, regenerated through the same ControlNet branch with an all-black
conditioning image. \ctrlglyph{}: conditioning on the closed vocabularies of
ten Chinese glyphs and ten Latin letters. \ctrlstr{}: the same type of content
under weak versus strong conditioning. \ctrlread{}: arbitrary open-vocabulary
strings.}
\label{fig:ctrl-examples}
\end{figure}

% =====================================================================
\section{Qualitative Deployment Examples}
\label{app:showcase}

Figure~\ref{fig:showcase} shows four production-derived \bench{} examples.
Given the transcription instruction, two frontier commercial MLLMs report
that no text is present, while \ours{}, still reading only the original
image, recovers the embedded promotional strings; a human viewer can read
the same strings when the page is viewed at reduced scale.

% Presentation follows the qualitative-table style of Vision-OPD:
% the image on top, model answers below a rule, our read-out highlighted.
\newcommand{\showcaseblock}[2]{%
  \begin{minipage}[t]{0.48\textwidth}
    \centering
    \includegraphics[height=4.6cm,keepaspectratio]{figures/#1}\\[3pt]
    \rule{\linewidth}{0.4pt}\\[4pt]
    {\small
    \begin{tabular}{@{}l@{\hskip 5pt}p{\dimexpr\linewidth-76pt\relax}@{}}
      User & What text is written in this image? \\[1pt]
      gemini-3.1-pro & ``There is no text in this image.'' \\[1pt]
      claude-opus-5 & ``There is no text in this image.'' \\[1pt]
      gpt-6-astra & ``There is no text in this image.'' \\[1pt]
      \ours{} & \textcolor{numblue}{\zh{#2}} \\
    \end{tabular}}
  \end{minipage}}

\begin{figure}[h]
\centering
\showcaseblock{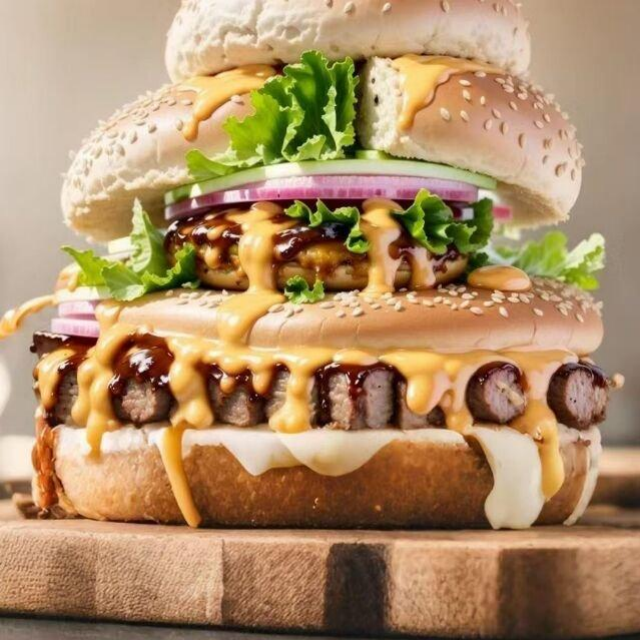}{浏览器搜78wm.cc}\hfill
\showcaseblock{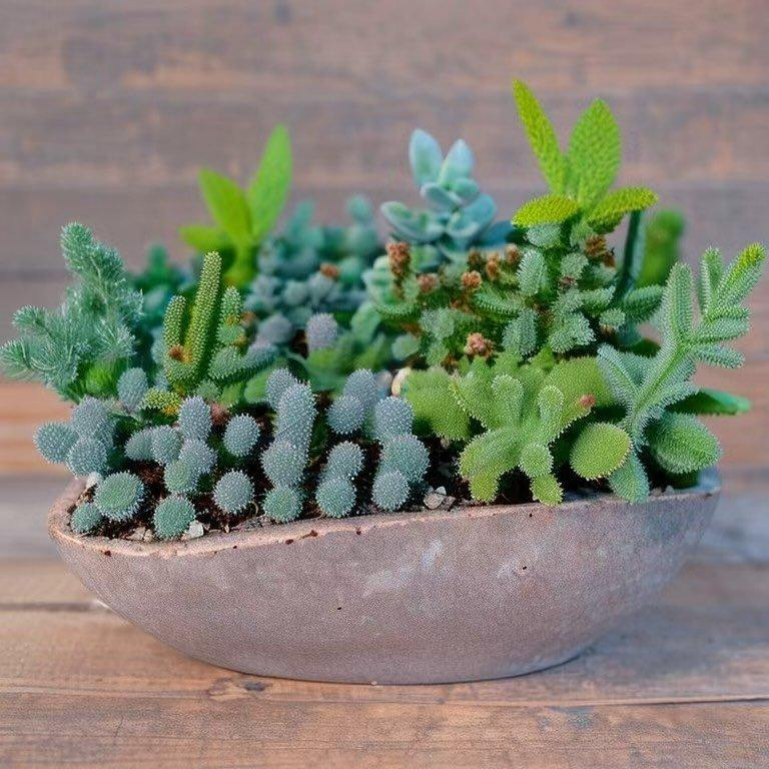}{浏览器搜58hx.cc}\\[10pt]
\showcaseblock{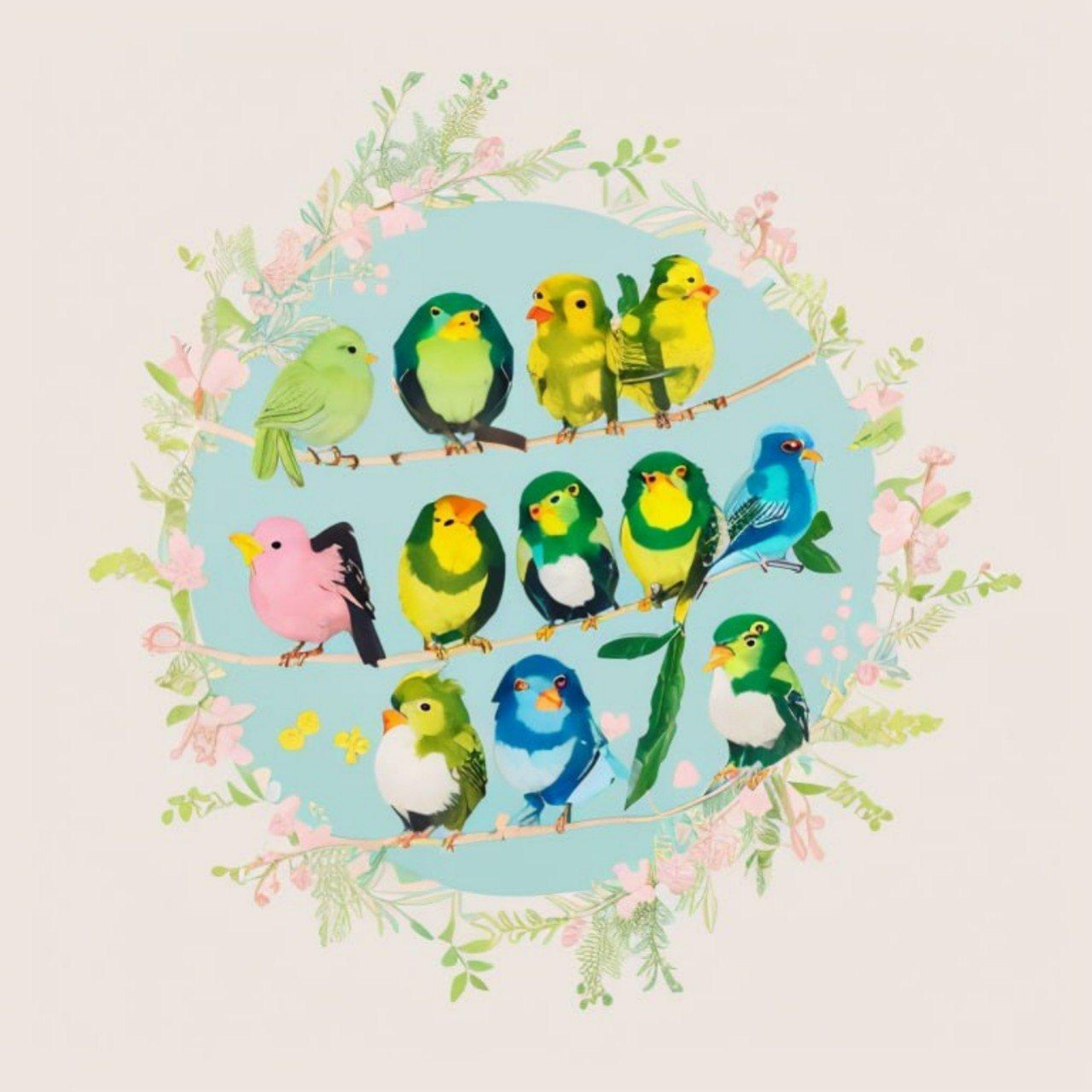}{Q群198943973}\hfill
\showcaseblock{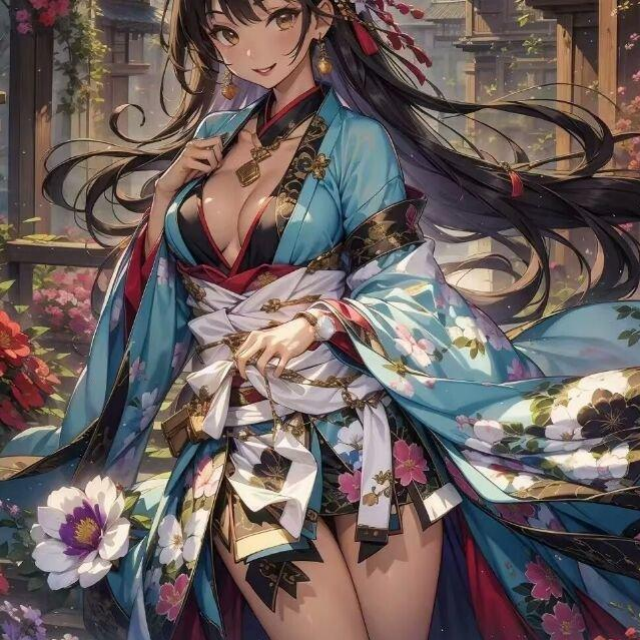}{搜索ov91.cc}
\caption{\textbf{\bench{} deployment examples that commercial models miss.}
Four production-derived moderation images, shown together with the answers of
two frontier commercial MLLMs and \ours{} to the transcription instruction.
Both commercial models report that no text is present on the original image,
whereas \ours{} reads out the embedded promotional strings; the same strings
become legible to a human when the page is squinted at or viewed at reduced
scale. These examples come from the production-derived portion of \bench{}
and contain no personal data.}
\label{fig:showcase}
\end{figure}

\subsection{\textcolor{blue}{Commercial-model failure breakdown}}
\label{app:commercial-failures}
\begingroup
\color{blue}
Table~\ref{tab:commercial-failures} breaks down commercial-model failures
under three input conditions: the original image, a view downscaled to
$1/8$ of the original linear resolution, and a $40{\times}40$ view.
The reported failures are partitioned into nonempty transcriptions with
ANLS~$=0$, responses explaining that the model cannot read the text or view
the image, empty answers, and other responses. The four category counts
sum to the total failures in each row. This breakdown distinguishes
unsuccessful transcription attempts from empty answers and explanatory
responses.

\par\medskip\noindent
\begin{minipage}{\textwidth}
\makeatletter\def\@captype{table}\makeatother
\centering\small
\setlength{\tabcolsep}{4pt}
\caption{\textcolor{blue}{Commercial-model failure categories on \bench{}.
Counts are reported for each model and input view. ``Nonempty, ANLS~$=0$''
denotes nonempty transcription attempts receiving zero ANLS;
``Explanation'' denotes cannot-read/view explanations.}}
\label{tab:commercial-failures}
\begin{tabular}{@{}llrrrrr@{}}
\toprule
Model & View & Total failures & \shortstack{Nonempty,\\ANLS $=0$} & Explanation & Empty & Other \\
\midrule
GPT-astra & Original & 947 & 365 & 561 & 19 & 2 \\
 & $1/8$ & 947 & 244 & 701 & 1 & 1 \\
 & $40{\times}40$ & 974 & 223 & 750 & 0 & 1 \\
\midrule
Claude & Original & 744 & 307 & 0 & 437 & 0 \\
 & $1/8$ & 745 & 227 & 59 & 459 & 0 \\
 & $40{\times}40$ & 881 & 607 & 112 & 162 & 0 \\
\midrule
Gemini & Original & 816 & 371 & 0 & 445 & 0 \\
 & $1/8$ & 668 & 190 & 0 & 478 & 0 \\
 & $40{\times}40$ & 350 & 329 & 0 & 21 & 0 \\
\bottomrule
\end{tabular}
\end{minipage}
\par\medskip
\endgroup

\section{\textcolor{blue}{Generalization to Additional Models}}
\label{app:model-generalization}
\begingroup
\color{blue}
We further compare Base, SFT, and \ours{} on InternVL 3.5-8B and Ministral 3-3B. Table~\ref{tab:model-generalization} reports their transcription results on \bench{}. On InternVL 3.5-8B, \ours{} improves over SFT by \num{12.50} points, extending the gain beyond the main backbone. On Ministral 3-3B, however, OPSD underperforms SFT by \num{35.60} points.

\begin{table}[h]
\centering\small
\caption{\textcolor{blue}{Generalization to additional models on \bench{}. Scores are ANLS (\%); $\Delta$ is OPSD minus SFT in percentage points.}}
\label{tab:model-generalization}
\begin{tabular}{@{}lrrrr@{}}
\toprule
Model & Base & SFT & OPSD & $\Delta$ \\
\midrule
InternVL 3.5-8B & \num{12.07} & \num{60.87} & \textbf{\num{73.37}} & $+12.50$ \\
Ministral 3-3B & \num{6.88} & \textbf{\num{48.11}} & \num{12.51} & $-35.60$ \\
\bottomrule
\end{tabular}
\end{table}

\paragraph{Teacher instruction following on Ministral 3-3B.}
In an inspection of training samples, the teacher achieves a mean ANLS of only \num{30.56} despite receiving the original image, the downscaled view, and the correct transcription. For example, even when the hint explicitly supplies \emph{Megan Kelley Hall}, the teacher instead generates a book description and a list of authors. This indicates unstable adherence to the answer hint in this setting: the teacher can provide off-task content rather than reliable transcription supervision. This weakness may contribute to the poor distillation result. The contrasting outcomes show that the benefit of OPSD is not uniform across models and depends on the quality of the teacher supervision.
\endgroup

\section{\textcolor{blue}{SquintBench Coverage and Data Access}}
\label{app:bench-coverage}
\begingroup
\color{blue}
The synthetic diagnostic images are publicly available at
\url{https://anonymous-hf.com/a/jci2bjb3m7x2/}.
The training data are available at
\url{https://anonymous-hf.com/a/qjh55c6odaz5/}.

Figure~\ref{fig:bench-diversity} presents twelve examples selected from the
\num{1076}-image benchmark to illustrate variation in visual style, scene
content, and text arrangement. The selection includes anime illustrations,
photorealistic portraits, cyberpunk scenes, fantasy characters, neon street
scenes, graphic posters, and object-based compositions, with single-line,
multi-line, and diagonal text arrangements.
The examples span several collection-source groups recorded in the benchmark
metadata. This is a qualitative, deliberately varied selection rather than
an estimate of the prevalence of each style in the benchmark.

\begin{figure}[htbp]
\centering
\includegraphics[width=\textwidth]{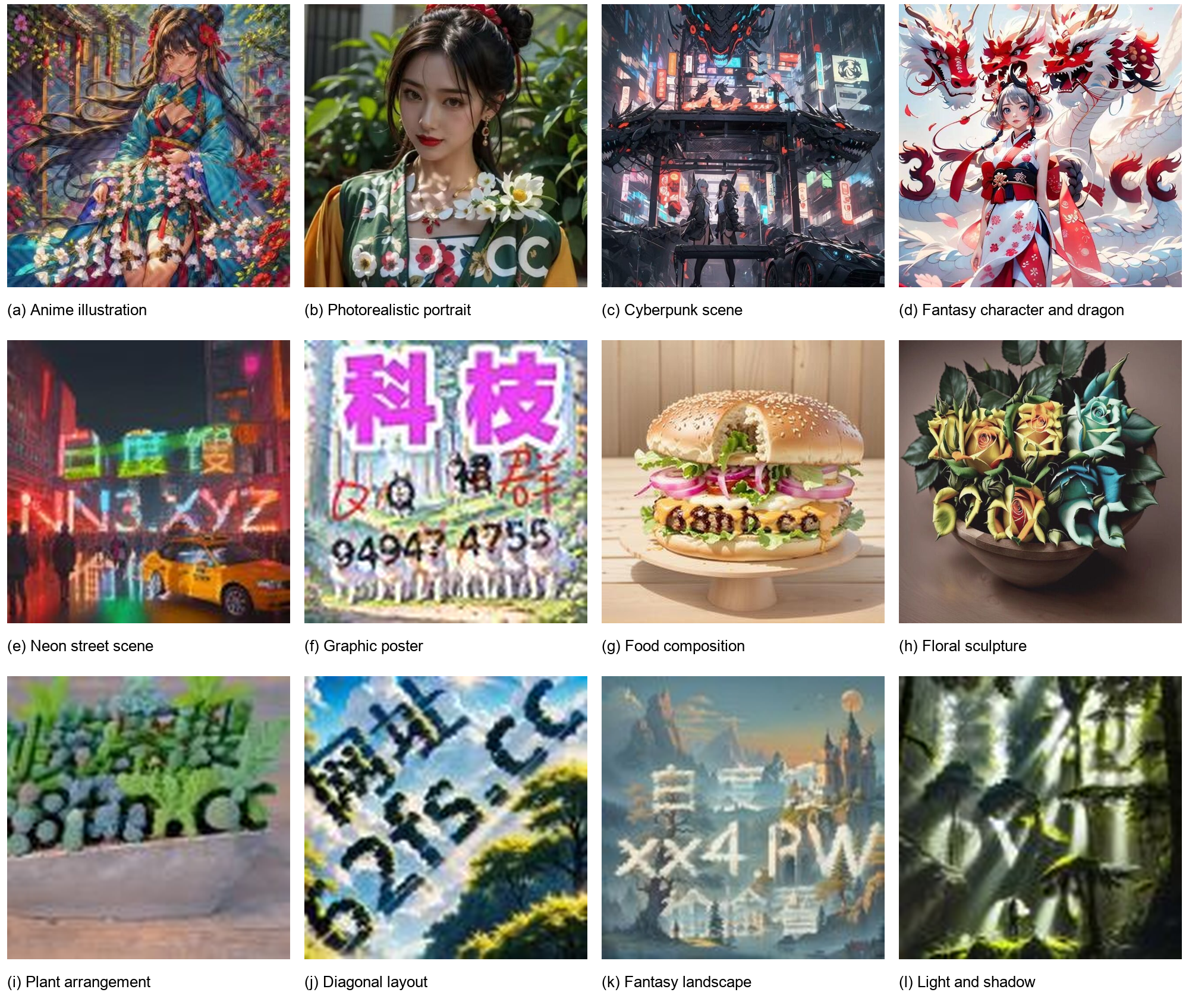}
\caption{\textcolor{blue}{\textbf{Visual diversity in SquintBench.} Twelve production-derived
examples illustrate different visual styles and ways of embedding text in
scene structure, object arrangements, textures, and shading. Original image
content and aspect ratios are preserved; images are resized for display.
These examples illustrate benchmark coverage rather than establish
representativeness of all squint-text images.}}
\label{fig:bench-diversity}
\end{figure}
\endgroup

\end{document}